\documentclass[lettersize,journal]{IEEEtran}

\usepackage{amsmath,amssymb,amsfonts}
\usepackage{mathrsfs}
\usepackage{bm}
\usepackage{bbm}
\usepackage{dsfont}

\usepackage{graphicx}
\usepackage{booktabs}
\usepackage{multirow}
\usepackage{array}

\usepackage[caption=false,font=normalsize,labelfont=sf,textfont=sf]{subfig}

\usepackage{algorithm}
\usepackage{algorithmic}

\usepackage{algorithm}
\usepackage{algorithmic}

\usepackage{stfloats}
\usepackage{setspace}

\usepackage{textcomp}
\usepackage{url}
\usepackage{verbatim}
\usepackage{cite}
\usepackage{color}

\usepackage{lineno}
\usepackage{hyperref}

\usepackage{times}
\usepackage{mathptmx}

\usepackage{makecell}
\usepackage{booktabs}
\usepackage{multirow}
\usepackage{colortbl}

\definecolor{lightred}{RGB}{255,226,226}
\definecolor{lightblue}{RGB}{224, 236, 255}
\definecolor{lightgreen}{RGB}{228, 244, 226}

\begin{document}

\title{CRT-HMAR: Causal Requirement Tracing-Guided Hierarchical Multi-Agent Regulation for Open-Task-Aware Infrared-Visible Image Fusion}

\author{Zengyi Yang, Shuai Yuan, Zhong-Cheng Wu, Juan Cheng, Huafeng Li, and Yu Liu,~\IEEEmembership{Senior Member, IEEE}

\thanks{This work was supported in part by the Fundamental and Interdisciplinary Disciplines Breakthrough Plan of the Ministry of Education of China under Grant JYB2025XDXM109, and in part by the National Natural Science Foundation of China under Grant 62576132.}
\thanks{Z. Yang, S. Yuan, Z. Wu, J. Cheng, and Y. Liu are with the School of Instrument Science and Opto-electronics Engineering, Hefei University of Technology, Hefei 230009, China (e-mail: zengyiyang0211@mail.hfut.edu.cn; shuaiyuan@hfut.edu.cn; zhongcheng.wu@hfut.edu.cn; chengjuan@hfut.edu.cn; yuliu@hfut.edu.cn).}

\thanks{H. Li is with the Faculty of Information
	Engineering and Automation, Kunming University of Science and
	Technology, Kunming 650500, China (e-mail: lhfchina99@kust.edu.cn).}

}
	


\maketitle

\begin{abstract}
Infrared and visible (IR-VIS) image fusion integrates complementary multimodal information into a single fused image to support downstream vision tasks. However, existing methods are typically tailored to seen tasks within a fixed task set and struggle to generalize to unseen tasks, which restricts their applicability in real-world open-task scenarios. To address this issue, this paper proposes CRT-HMAR, a Causal Requirement Tracing-Guided Hierarchical Multi-Agent Regulation Framework for open-task-aware IR-VIS image fusion. CRT-HMAR introduces a Causal Requirement Tracing Task Localization mechanism, which actively intervenes in key image information and observes task-network response variations to map task-specific semantic preferences into image-level causal requirement maps. Based on these maps, a requirement analysis agent aggregates task-specific requirement knowledge to adaptively guide requirement-customized image fusion. Moreover, CRT-HMAR incorporates History-Analysis Multi-Objective Balancing and Task-Level-Correction Conflict Mitigation mechanisms, jointly constructing a hierarchical regulation chain of “requirement interpretation $ \to $ task balancing $ \to $ conflict mitigation”. Through multiple collaborative agents, CRT-HMAR dynamically regulates key processes including open-task requirement modeling, multi-task balanced optimization, and gradient conflict mitigation. Extensive experiments on open-task scenarios involving five downstream tasks demonstrate that CRT-HMAR significantly improves generalization to unseen tasks while maintaining the performance and balance of seen tasks. Overall, CRT-HMAR shifts IR-VIS image fusion from task-oriented modeling toward requirement-oriented modeling, promoting its extension from closed-task settings to real-world open-task scenarios.
\end{abstract}

\begin{IEEEkeywords}
Infrared and visible image fusion, open-task-aware image fusion, causal requirement tracing, hierarchical regulation chain.
\end{IEEEkeywords}

\section{Introduction}\label{sec1}

Infrared and visible (IR-VIS) image fusion \cite{55, 121, 122, 120} aims to effectively integrate multimodal images captured by infrared and visible sensors, thereby fusing complementary information from different modalities into a single image to provide a more comprehensive description of scene content. Specifically, infrared images contain prominent thermal radiation information, which can effectively highlight objects such as pedestrians and vehicles in the scene, while visible images provide rich texture details, offering fine-grained semantic descriptions of the objects. Benefiting from the complementary advantages of multimodal information in fused images, IR-VIS image fusion has been widely employed to improve the performance of high-level vision tasks, such as object detection, semantic segmentation, salient object detection, pedestrian detection, and depth estimation, and has demonstrated significant application value in fields such as aerospace and autonomous driving.

\begin{figure}
	\begin{center}
		\includegraphics[width=1\linewidth]{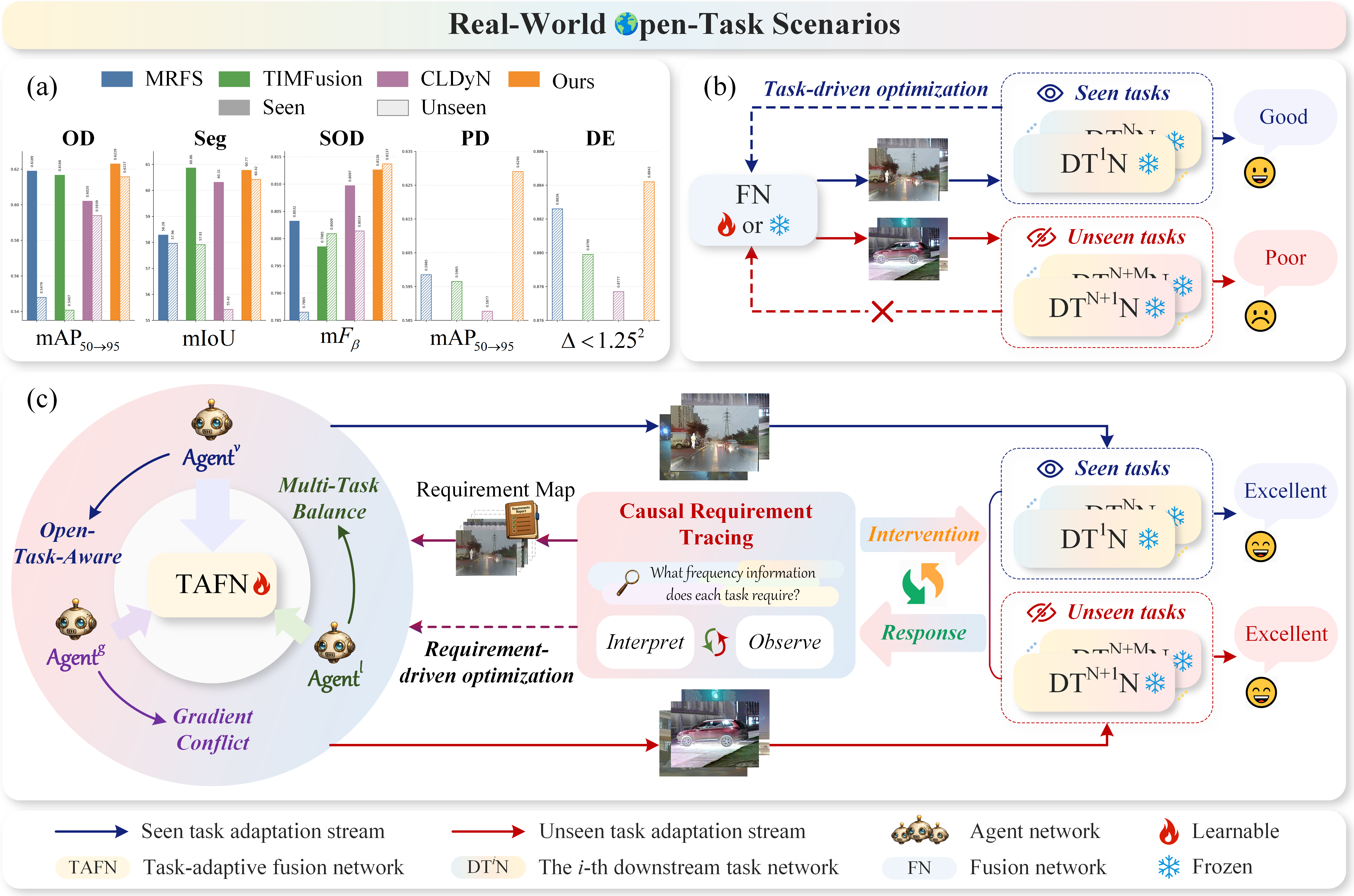}
	\end{center}
	\caption{Comparison between the proposed method and existing methods. (a) Performance comparison with existing methods in open-task scenarios. (b) Existing task-oriented modeling paradigm. (c) The proposed method adopts a Causal Requirement Tracing-guided Hierarchical multi-Agent Regulation framework and shifts toward a requirement-oriented modeling paradigm.}
	\label{fig1}
\end{figure}

Therefore, task-aware IR-VIS image fusion methods that aim to improve the performance of downstream high-level vision tasks have attracted increasing attention in recent years. According to the range of tasks they adapt to, existing methods can be roughly divided into two categories: single-task-aware IR-VIS image fusion methods and multi-task-aware IR-VIS image fusion methods. Among them, single-task-aware IR-VIS image fusion methods can be further categorized into task semantic supervision methods and task semantic injection methods, depending on whether the semantic information from downstream task networks is explicitly exploited. The former typically cascade the fusion network with a specific downstream task network and optimize the fusion network using task-related losses to adapt to specific task requirements, with representative methods including SeAFusion \cite{3}, TarDAL \cite{8}, and TIMFusion \cite{65}. The latter further introduce semantic features from downstream task networks into the fusion process, enabling the fused image to contain task-related semantic information, with representative methods including MRFS \cite{90}, SAGE \cite{73}, and UAAFusion \cite{72}. Although the above methods can effectively improve the performance of specific tasks, their adaptation objectives are mainly oriented toward single-task scenarios, making it difficult for them to simultaneously satisfy the requirements of multiple downstream tasks.

To further enhance the adaptability of IR-VIS image fusion in complex multi-task scenarios, multi-task-aware IR-VIS image fusion methods have gradually received attention in recent years, with representative methods including IDF-TDDT \cite{70} and CLDyN \cite{125}. Specifically, IDF-TDDT modulates multimodal features in the fusion network using multiple task-related instructions, enabling the model to simultaneously adapt to the requirements of multiple downstream tasks. CLDyN proposes a closed-loop optimization mechanism that dynamically customizes the network structure and network weights using semantic features fed back from downstream task networks, thereby enabling fine-grained task adaptation and improving the multi-task adaptability of the fusion model.

However, as shown in Fig. \ref{fig1}(a), although existing task-aware IR-VIS image fusion methods can improve the performance of downstream tasks within a fixed task set in either single-task or multi-task scenarios, they struggle to effectively adapt to unseen downstream tasks beyond the task set, and thus remain limited in real-world open-task scenarios. Specifically, the representative methods shown in the figure include single-task-aware methods, i.e., TIMFusion as a task semantic supervision method and MRFS as a task semantic injection method, as well as the multi-task-aware method CLDyN. When downstream tasks shift from seen to unseen ones, these methods usually suffer from significant performance degradation. Moreover, they exhibit poor performance on unseen tasks outside the predefined task set. This is mainly because existing methods generally follow a task-oriented modeling paradigm, in which the training process of the fusion model is directly driven by specific downstream tasks, as shown in Fig. \ref{fig1}(b). Under this paradigm, task feedback or optimization objectives directly act on the fusion network, making the model prone to being highly coupled with the seen tasks, thereby limiting its generalization ability to unseen tasks.

To address this issue, as shown in Fig. \ref{fig1}(c), this paper proposes a Causal Requirement Tracing-Guided Hierarchical Multi-Agent Regulation Framework (CRT-HMAR) for open-task-aware IR-VIS image fusion. Different from existing task-oriented modeling paradigms, CRT-HMAR shifts toward requirement-oriented modeling, where the fusion model is driven by explicitly characterized task requirements rather than specific tasks, thereby reducing its coupling with seen tasks and improving its generalization to unseen tasks in open-task scenarios. Specifically, a Causal Requirement Tracing Task Localization mechanism is designed to actively intervene in key information within fused images and observes the response variations of task networks, explicitly mapping the high-level semantic preferences implicit in downstream tasks into image-level causal requirement maps. Based on these maps, a requirement analysis agent interprets task requirements and adaptively aggregates task-specific requirement knowledge vectors, which are then used to guide the Task-Adaptive Fusion Network (TAFN) in performing semantic restoration on multimodal features and generating requirement-customized fused images. In addition, History-Analysis Multi-Objective Balancing and Task-Level-Correction Conflict Mitigation mechanisms are introduced to jointly construct a hierarchical regulation chain of “requirement interpretation $\to$ task balancing $\to$ conflict mitigation”, enabling dynamic regulation of open-task modeling and multi-task optimization. Specifically, the former leverages a task-balancing agent to analyze task performance variations and imbalance states, and dynamically adjusts the multi-task joint optimization objective to alleviate task imbalance. The latter introduces a conflict-mitigation agent to model the conflict and synergy between task-specific objectives and the joint objective, and corrects the joint gradient using task-specific gradients to mitigate gradient conflicts. In summary, the main contributions of this paper are as follows:

\begin{itemize}

\item \textbf{New Problem Formulation}: To the best of our knowledge, this is the first work to explore IR-VIS image fusion in open-task scenarios, promoting the extension of IR-VIS image fusion \textbf{\textit{from predefined closed-task scenarios to real-world open-task scenarios}}. To this end, this paper proposes CRT-HMAR, a Causal Requirement Tracing-Guided Hierarchical Multi-Agent Regulation framework. CRT-HMAR integrates Causal Requirement Tracing Task Localization, History-Analysis Multi-Objective Balancing, and Task-Level-Correction Conflict Mitigation mechanisms to jointly model requirement interpretation, task balancing, and conflict mitigation, thereby achieving adaptive fusion in open-task scenarios.

\item \textbf{Requirement-Oriented Modeling Paradigm}: This paper proposes a Causal Requirement Tracing Task Localization mechanism to explicitly transform the high-level semantic preferences implicit in downstream task networks into image-level requirement maps. This mechanism actively intervenes key information in the fused image and analyzes the response variations of downstream task networks, constructing a causal requirement map for characterizing task semantic requirements. On this basis, the agent adaptively aggregates task-specific requirement knowledge to guide requirement-customized image fusion. This process realizes the transition \textbf{\textit{from the existing task-oriented modeling paradigm to a requirement-oriented modeling paradigm}}, thus enhancing the adaptability of the model to both seen and unseen tasks.

\item \textbf{Hierarchical Regulation Chain}: This paper further introduces History-Analysis Multi-Objective Balancing and Task-Level-Correction Conflict Mitigation mechanisms to jointly construct a \textbf{\textit{hierarchical regulation chain of “requirement interpretation $ \to $ task balancing $ \to $ conflict mitigation”}}. Specifically, multiple collaborative agents are introduced to dynamically regulate key processes of requirement interpretation, multi-task balancing, and gradient conflict mitigation. This design further improves the generalization ability of the model to unseen tasks while maintaining multi-task balance among seen tasks.

\item \textbf{Strong Open-Task Adaptability}: Extensive experiments on diverse open-task scenarios involving five downstream tasks, including object detection, semantic segmentation, salient object detection, pedestrian detection, and depth estimation, demonstrate that, compared with existing task-aware IR-VIS image fusion methods, the proposed method exhibits more \textbf{\textit{competitive performance in terms of task generalization ability and open-task adaptability}}.

\end{itemize}

The rest of this paper is organized as follows: Section \ref{sec2} provides a comprehensive review of related work. Section \ref{sec3} elaborates on the proposed method in detail. Section \ref{sec4} presents the experimental results and effectiveness analysis. Finally, Section \ref{sec5} concludes this paper.

\section{Related Work}\label{sec2}

Existing IR-VIS image fusion techniques can be roughly divided into task-agnostic IR-VIS image fusion methods and task-aware IR-VIS image fusion methods, depending on whether their optimization objectives explicitly consider the requirements of downstream vision tasks.

\subsection{Task-Agnostic IR-VIS Image Fusion}

Task-agnostic IR-VIS image fusion methods aim to exploit the complementary information between multimodal images and integrate it into a single fused image, thereby improving the quality of the fused image to assist human decision-making. Early studies mainly relied on manually designed feature extraction mechanisms and achieved a large number of pioneering research results. These methods are referred to as Traditional IR-VIS image fusion methods, which typically employ techniques such as discrete wavelet transform (DWT) \cite{21}, ridgelet transform \cite{24}, non-subsampled contourlet transform (NSCT) \cite{20}, and sparse representation \cite{23} to decompose, extract, and reconstruct features from different modalities, generating high-fidelity fusion results. 

With the rapid development of deep learning techniques, recent deep learning-based IR-VIS image fusion methods typically exploit advanced network architectures to extract multimodal features and learn complex mappings between multi-source images and fused images from large-scale data. According to the adopted network architectures, these methods can be roughly categorized into convolutional neural network (CNN)-based methods \cite{28, 128, 29, 132}, Transformer-based methods \cite{31, 126, 129, 130}, Mamba-based methods \cite{123}, and hybrid architecture-based methods \cite{36, 131}. However, real-world complex scenarios are often affected by various degradation factors, which pose greater challenges to image fusion. To improve the applicability of fusion models in complex scenarios, IR-VIS image fusion methods for complex scenarios have been proposed. These methods aim to alleviate the impact of degradation factors, such as inaccurate registration between multimodal images \cite{10, 14, 82, 83, 84}, inconsistent spatial resolutions \cite{1, 2}, and extreme conditions \cite{94, 95, 96, 97, 99, 100, 117}, on image quality, thereby generating more stable and robust fused images. Although the above studies have significantly expanded the application boundaries of IR-VIS image fusion, their optimization objectives still mainly focus on the visual quality of the fused image itself, with insufficient attention paid to the practical requirements of fusion results in downstream high-level vision tasks. Consequently, their performance improvement in task-driven application scenarios remains limited.

\subsection{Task-Aware IR-VIS Image Fusion}

Task-aware IR-VIS image fusion methods aim to enhance the semantic information required by downstream tasks in fused images, thereby improving the performance of downstream tasks. According to the range of tasks to which the fusion network adapts, existing methods can be roughly divided into single-task-aware IR-VIS image fusion methods and multi-task-aware IR-VIS image fusion methods. 

Single-task-aware IR-VIS image fusion methods are typically optimized for a specific downstream task, assuming that the fusion network only needs to adapt to a single-task scenario. Depending on whether semantic information from task networks is explicitly introduced into the fusion process, these methods can be further divided into task semantic supervision methods and task semantic injection methods. Among them, task semantic supervision methods typically cascade the fusion network with a task network and optimize the fusion network using task-related losses. Representative methods, such as TarDAL \cite{8}, SeAFusion \cite{3}, and IRFS \cite{88}, employ losses related to object detection, semantic segmentation, and salient object detection, respectively, to guide the optimization of the fusion model. In contrast, MetaFusion \cite{7} introduces a semantic consistency constraint between fused features and semantic features of the task network. Beyond direct task supervision, CAF \cite{74} and TDFusion \cite{71} exploit task-related losses to dynamically optimize key parameters in fusion losses, thereby constructing task-customized fusion objectives. To balance visual quality and downstream task performance, BDLFusion \cite{104} and DCEvo \cite{105} further adjust the weighting coefficients between fusion losses and task losses. In addition, considering real-world degradation factors, TIMFusion \cite{65} employs a task-oriented network architecture search and parameter initialization strategy to improve task-adaptation robustness, while PAIF \cite{106} enhances noise resistance by constructing complex noisy training data.

Although task semantic supervision methods improve task performance through task losses, such indirect constraints remain limited in capturing task-related semantics. To address this limitation, task semantic injection methods introduce semantic information from task networks into the fusion process. For example, DetFusion \cite{9} uses attention maps fed back from the task network to aggregate multimodal features, while UAAFusion \cite{72} statistically analyzes pixel-wise contributions to task performance and constructs attention maps to guide fusion. However, such attention-based strategies mainly reaggregate semantic information, making it difficult to enrich high-level semantic representations in fused features. To inject semantics more directly, SegMiF \cite{6} designs an interaction mechanism between fused features and task-network semantic features. Considering that the semantic extraction capability of task networks trained on specific datasets is limited, SAGE \cite{73} leverages robust semantic features from the Segment Anything Model (SAM) \cite{108} to enhance task adaptability across scenarios. Nevertheless, a semantic gap usually exists between high-level task semantics and low-level image representations, which may hinder the effective use of injected semantics. To alleviate this issue, PSFusion \cite{89}, MRFS \cite{90}, and SDCFusion \cite{91} share partial features between the fusion network and the task network, enabling progressive transfer of high-level semantic information to the fusion results.

\begin{figure*}
	\begin{center}
		\includegraphics[width=0.96\linewidth]{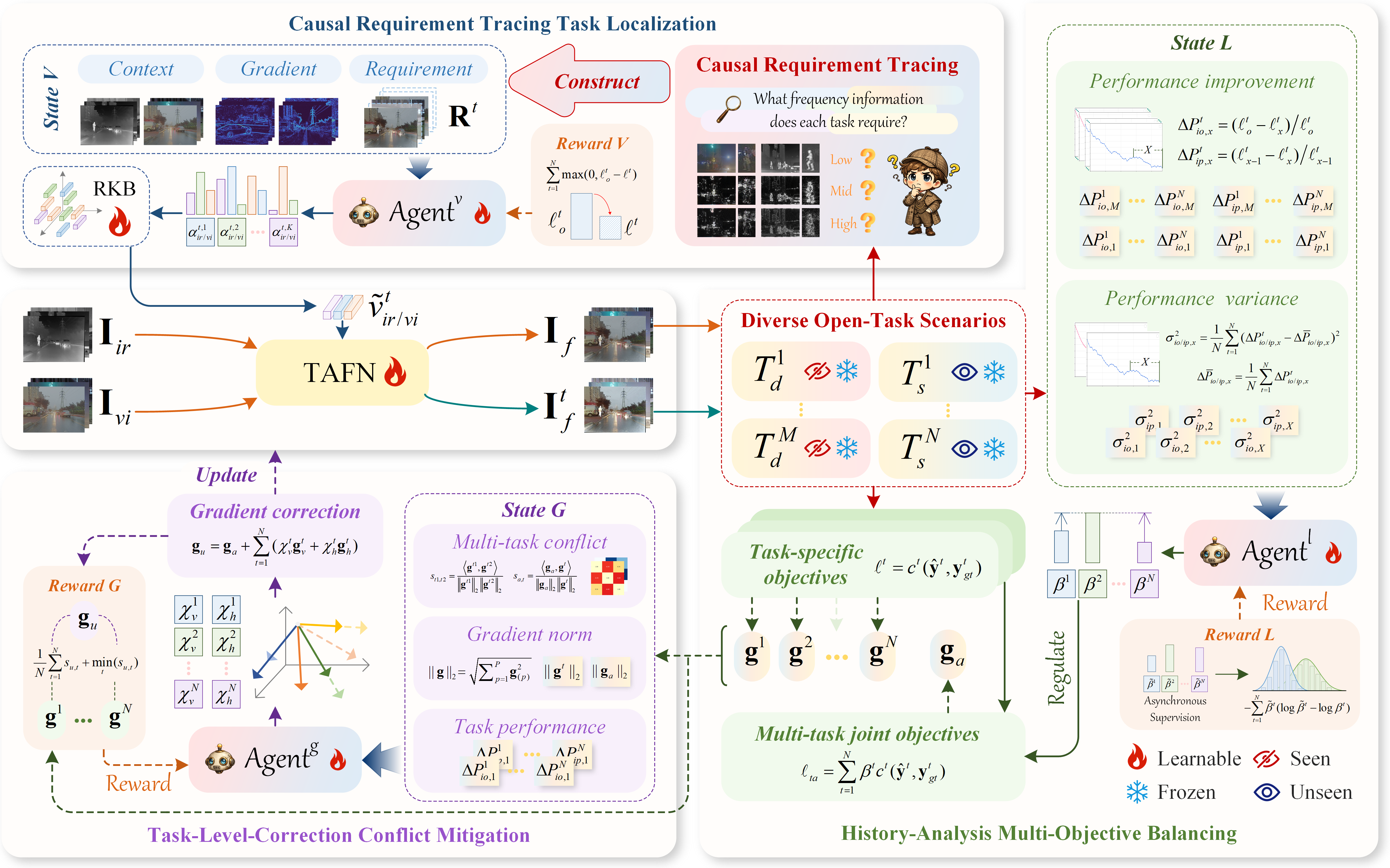}
	\end{center}
	\caption{
	Overview of the proposed CRT-HMAR. CRT-HMAR constructs causal requirement maps through the Causal Requirement Tracing Task Localization mechanism. ${\rm{Agent}^v}$ is introduced to interpret these maps and aggregate requirement knowledge vectors from the RKB. Guided by these vectors, the TAFN generates requirement-customized fused images. In addition, the History-Analysis Multi-Objective Balancing and Task-Level-Correction Conflict Mitigation mechanisms employ ${\rm{Agent}^{l/g}}$, respectively, to dynamically regulate multi-task optimization, thereby alleviating multi-task imbalance and gradient conflicts.
	}
	\label{fig2}
\end{figure*}

Although the above methods can improve performance on specific downstream tasks, their optimization is usually centered on a single task, making them difficult to extend to multi-task scenarios. To address this limitation, multi-task-aware IR-VIS image fusion methods have recently attracted increasing attention. Representative methods, such as IDF-TDDT \cite{70} and CLDyN \cite{125}, aim to enable a single fusion network to adapt to multiple downstream tasks simultaneously. Specifically, IDF-TDDT dynamically modulates multimodal features in the fusion network using different downstream task-related instructions provided by users, enabling the fusion network to satisfy the requirements of multiple downstream tasks. CLDyN proposes a closed-loop optimization mechanism that customizes the network structure and network weights by leveraging semantic features fed back from different downstream task networks, thereby enabling the fusion model to dynamically adapt to diverse task requirements. Although these methods can adapt to seen downstream tasks within a fixed task set that participate in training, they suffer from significant performance degradation when deployed on unseen downstream tasks beyond the task set, which limits their application in broader real-world open-task scenarios.

To improve generalization to unseen tasks, this paper proposes a Causal Requirement Tracing-guided Hierarchical multi-Agent Regulation framework (CRT-HMAR) for open-task-aware IR-VIS image fusion. CRT-HMAR introduces a Causal Requirement Tracing Task Localization mechanism, which actively intervenes in key information within fused images and observes task network response variations to interpret the high-level semantic preferences implicit in downstream tasks. Furthermore, History-Analysis Multi-Objective Balancing and Task-Level-Correction Conflict Mitigation mechanisms construct a hierarchical regulation chain of “requirement interpretation $\to$ task balancing $\to$ conflict mitigation”. Through multiple collaborative agents, this regulation chain dynamically coordinates open-task requirement modeling and multi-task optimization. With these mechanisms, CRT-HMAR achieves better adaptability to real-world open-task scenarios.

\section{The Proposed Method}\label{sec3}

\subsection{Overview}\label{sec3_0}

Let the set of downstream tasks used for model training be denoted as ${T_s} = \{ T_s^1,...,T_s^N\}$, and the set of downstream tasks used for model evaluation be denoted as ${T_d} = \{ T_d^1,...,T_d^M\} $, where ${T_s} \cap {T_d} = \emptyset $. The main objective of this paper is to enable the model pretrained on ${T_s}$ to simultaneously adapt to ${T_s} \cup {T_d}$ without any parameter tuning. To this end, this paper proposes CRT-HMAR, a Causal Requirement Tracing-Guided Hierarchical Multi-Agent Regulation Framework for open-task-aware IR-VIS image fusion.

As shown in Fig. \ref{fig2}, the proposed method constructs a hierarchical regulation chain of “requirement interpretation $\to$ task balancing $\to$ conflict mitigation” through three mechanisms: Causal Requirement Tracing Task Localization, History-Analysis Multi-Objective Balancing, and Task-Level-Correction Conflict Mitigation, for adapting the fusion model to open-task scenarios. Specifically, the Causal Requirement Tracing Task Localization mechanism actively intervenes in key information within fused images and observes the responses of downstream task networks, thereby interpreting the implicit semantic preferences of specific tasks and constructing causal requirement maps. Based on these maps, the requirement interpretation agent adaptively aggregates requirement knowledge vectors from the Requirement Knowledge Base (RKB), thereby guiding the Task-Adaptive Fusion Network (TAFN) to generate requirement-customized fused images. The History-Analysis Multi-Objective Balancing mechanism introduces a task balancing agent to monitor performance gains and imbalance states across tasks, enabling dynamic regulation of the multi-task loss function to suppress task imbalance. The Task-Level-Correction Conflict Mitigation mechanism introduces a conflict mitigation agent to analyze gradient conflicts between task-specific objectives and the multi-task joint objective, and further corrects the joint gradient using task-specific gradients to alleviate multi-task gradient conflicts.


\subsection{Task-Adaptive Fusion Network}\label{sec3_1}

\begin{figure}
	\begin{center}
		\includegraphics[width=0.96\linewidth]{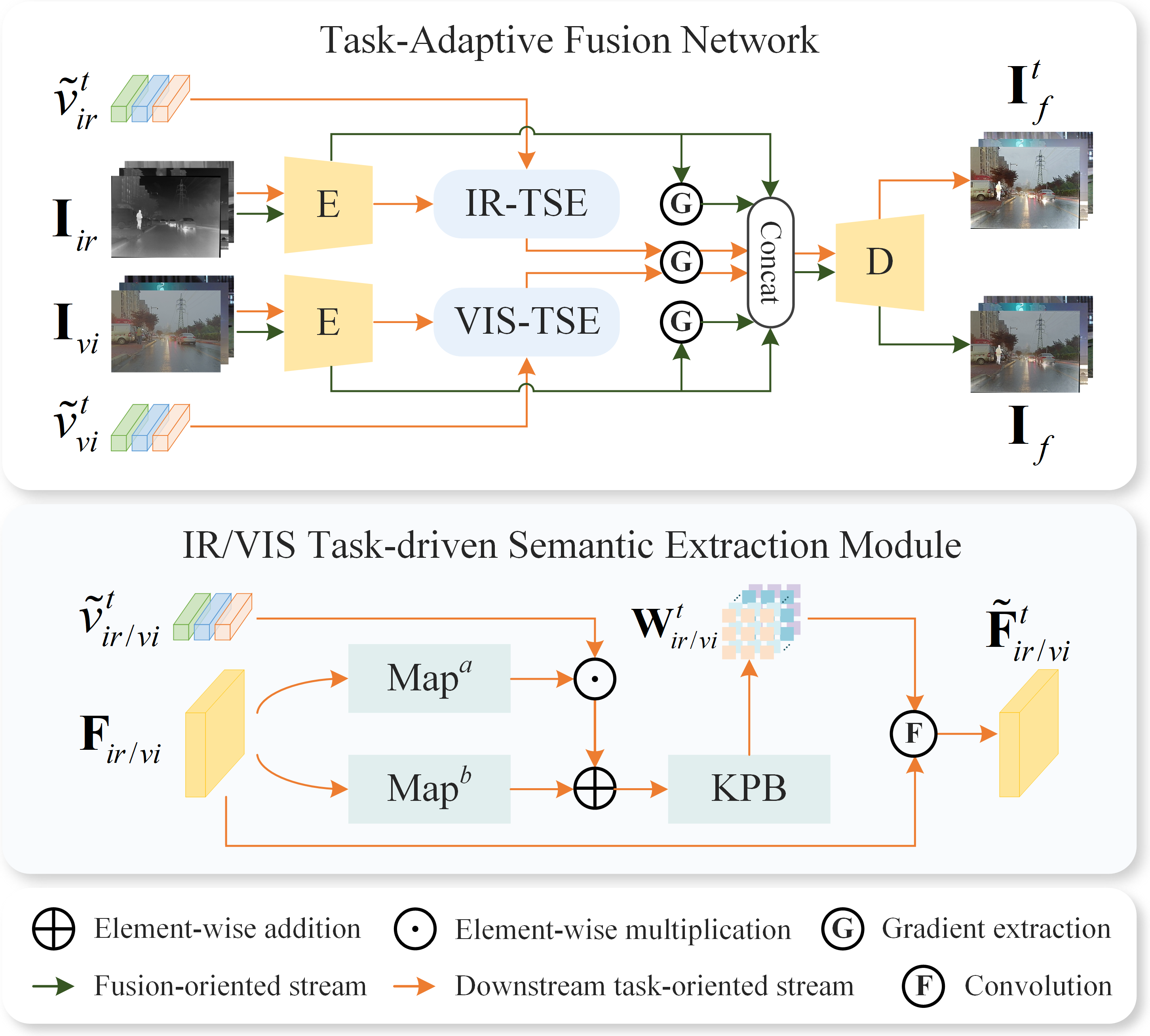}
	\end{center}
	\caption{Network architecture of TAFN. TAFN performs semantic restoration on IR-VIS features according to task-specific requirement knowledge vectors, thereby reconstructing requirement-customized fused images.}
	\label{fig3}
\end{figure}

The Task-Adaptive Fusion Network (TAFN) is used to fuse IR-VIS images to obtain fused images. When oriented toward the image fusion task, TAFN directly generates visually dominated fused images from IR-VIS images. When oriented toward a specific downstream task, TAFN performs semantic restoration on IR-VIS features according to the requirement knowledge vectors, thereby reconstructing fused images that satisfy the requirements of the specific task. As shown in Fig. \ref{fig3}, TAFN mainly consists of an encoder, IR/VIS Task-driven Semantic Extraction (IR/VIS-TSE) Modules, and a decoder. The encoder is used to extract features. IR/VIS-TSE generates dynamic convolution kernels according to the requirement knowledge vectors to extract semantic features specified by the requirements. The decoder is used to reconstruct the fused image.

Specifically, when oriented toward the image fusion task, we feed the IR-VIS images ${{\bf{I}}_{ir/vi}}$ into the encoder (E), respectively, to obtain the  ${{\bf{F}}_{ir/vi}}$. Then, ${{\bf{F}}_{ir/vi}}$ and their gradient maps are concatenated and fed into the decoder (D) to obtain the fused image ${{\bf{I}}_f}$:
\begin{equation}
	\begin{aligned}
	{{\bf{I}}_f} = D([E({{\bf{I}}_{ir}}),\nabla E({{\bf{I}}_{ir}}),E({{\bf{I}}_{vi}}),\nabla E({{\bf{I}}_{vi}})]),		
	\end{aligned}
\end{equation}
where $\left[  \cdot  \right]$ denotes the concatenation operation along the channel dimension, and $\nabla $ denotes the Sobel gradient extraction operator. The encoder and decoder are mainly composed of two convolutional layers and three convolutional layers, respectively. To encourage ${{\bf{I}}_f}$ to contain as much multimodal information as possible, we introduce the fusion loss ${\ell _f}$:
\begin{equation}
	\begin{aligned}
		{\ell _f} = {\left\| {\nabla {{\bf{I}}_f} - \max (\nabla {{\bf{I}}_{ir}},\nabla {{\bf{I}}_{vi}})} \right\|_1} + \lambda {\left\| {{{\bf{I}}_f} - \max ({{\bf{I}}_{ir}},{{\bf{I}}_{vi}})} \right\|_1},
	\end{aligned}
\end{equation}
where $\lambda$ is the weighting coefficient for balancing the loss terms, which is set to 0.2 in this paper.

When oriented toward a specific downstream task, we feed ${{\bf{F}}_{ir/vi}}$ together with the requirement knowledge vector $\tilde v_{ir/vi}^t \in {^{1 \times e}}$ of the $t$-th task into the IR/VIS-TSE module. The $\tilde v_{ir/vi}^t$ is generated by the Causal Requirement Tracing Task Localization mechanism, as detailed in Section \ref{sec3_2}. To enable $\tilde v_{ir/vi}^t$ to adapt to diverse scene variations, we feed ${{\bf{F}}_{ir/vi}}$ into ${\rm{Ma}}{{\rm{p}}^a}$ and ${\rm{Ma}}{{\rm{p}}^b}$, respectively. The obtained scene features are respectively combined with $\tilde v_{ir/vi}^t$ through element-wise multiplication and addition, and the resulting feature is then fed into the Kernel Prediction Block (KPB) to predict the dynamic convolution kernel ${\bf{W}}_{ir/vi}^t$:
\begin{equation}
	\begin{aligned}
		{\bf{W}}_{ir/vi}^t = {\rm{KPB(Ma}}{{\rm{p}}^a}{\rm{(}}{{\bf{F}}_{ir/vi}}{\rm{)}} \odot \tilde v_{ir/vi}^t{\rm{ + Ma}}{{\rm{p}}^b}{\rm{(}}{{\bf{F}}_{ir/vi}}{\rm{))}},
	\end{aligned}
\end{equation}
where $ \odot $ denotes the element-wise multiplication operation. ${\rm{Ma}}{{\rm{p}}^a}$ and ${\rm{Ma}}{{\rm{p}}^b}$ are mainly composed of two convolutional layers and one global average pooling layer, while KPB mainly consists of two linear mapping layers. We use ${\bf{W}}_{ir/vi}^t$ to extract the semantic features corresponding to the task requirements from ${{\bf{F}}_{ir/vi}}$, and then reinject the semantic features into ${{\bf{F}}_{ir/vi}}$, thereby completing the semantic restoration process dominated by task-specific requirements:
\begin{equation}
	\begin{aligned}
		{\bf{\tilde{F}}}_{ir/vi}^t = {\bf{W}}_{ir/vi}^t * {{\bf{F}}_{ir/vi}} + {{\bf{F}}_{ir/vi}},
	\end{aligned}
\end{equation}
where $ * $ denotes the convolution operation, and ${\bf{\tilde {F}}}_{ir/vi}^t$ denotes the IR/VIS features after semantic restoration. We concatenate ${\bf{\tilde {F}}}_{ir/vi}^t$ and its gradient map, and feed them into the decoder to obtain the fused image ${\bf{I}}_f^t$ required by the $t$-th task. To enable ${\bf{I}}_f^t$ to better improve downstream task performance, we introduce the task-aware loss ${\ell _{ta}}$:
\begin{equation}
	\begin{aligned}
		{\ell _{ta}} = \sum\limits_{t = 1}^N {{\beta ^t}{c^t}({{{\bf{\hat y}}}^t},{\bf{y}}_{gt}^t)},
	\end{aligned}
\end{equation}
where ${c^t}( \cdot )$ denotes the loss function of the $t$-th task, ${{\bf{\hat y}}^t}$ denotes the downstream task result after semantic restoration for the requirement of the $t$-th task, and ${\bf{y}}_{gt}^t$ denotes the ground truth (GT) of the $t$-th task. ${\beta ^t}$ denotes the weight used to balance the loss terms of multiple tasks, which is dynamically adjusted by the History-Analysis Multi-Objective Balancing mechanism, as detailed in Section \ref{sec3_3}.

\subsection{Causal Requirement Tracing Task Localization}\label{sec3_2}

The Causal Requirement Tracing Task Localization mechanism constructs a causal requirement map for task-specific requirements by actively intervening key information in the fused image and observing the responses of downstream task networks. Based on the constructed causal requirement map, the requirement analysis agent integrates knowledge vectors that serve task-specific requirements from the Requirement Knowledge Base (RKB), thereby supporting the generation of requirement-customized fused images. This mechanism mainly consists of the Causal Requirement Tracing (CRT) module, the ${\rm{Agent}}_{ir/vi}^v$, and the RKB. As shown in Fig. \ref{fig2}, the CRT module actively intervenes in different frequency information of the fused image and constructs a causal requirement map according to the responses of the task networks. ${\rm{Agent}}_{ir/vi}^v$ is used to interpret these map and accordingly integrate requirement knowledge vectors from the RKB. The RKB ${{\bf{V}}_{ir/vi}} \in {^{K \times e}}$ provides ${\rm{Agent}}_{ir/vi}^v$ with rich image-level requirement-related knowledge to support its requirement interpretation process.

\begin{figure}
	\begin{center}
		\includegraphics[width=0.96\linewidth]{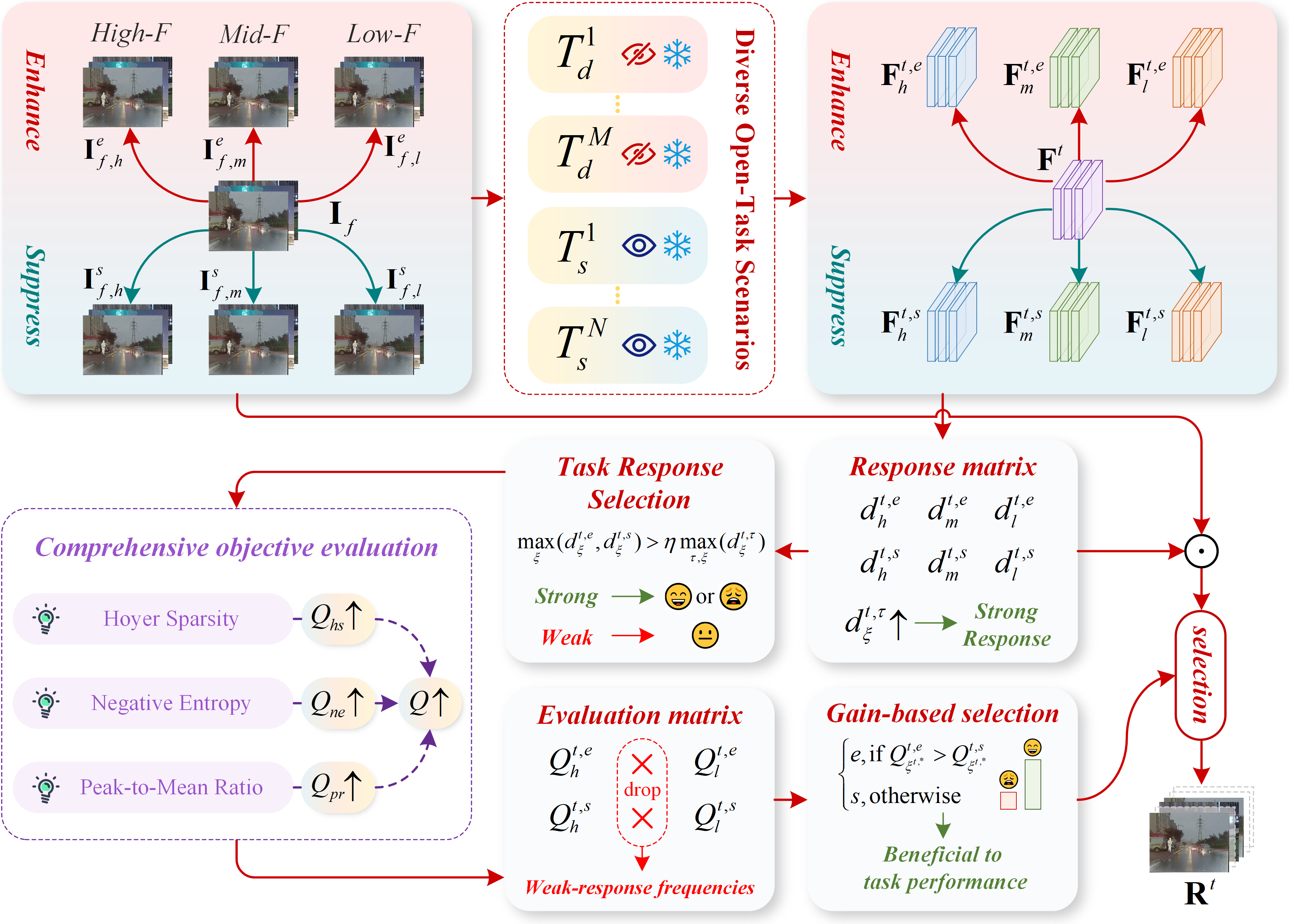}
	\end{center}
	\caption{Architecture of the CRT module. The CRT module actively intervenes in key information within the fused image and observes the response variations of downstream task networks, thereby constructing causal requirement maps.}
	\label{fig4}
\end{figure}

As shown in Fig. \ref{fig4}, we feed ${{\bf{I}}_f}$ into the CRT module, and decompose it into the high-frequency component ${{\bf{I}}_h}$, the middle-frequency component ${{\bf{I}}_m}$, and the low-frequency component ${{\bf{I}}_l}$ using two-dimensional fast Fourier transform combined with frequency-domain masking, where ${{\bf{I}}_f} = {{\bf{I}}_h} + {{\bf{I}}_m} + {{\bf{I}}_l}$. We then actively enhance or suppress the high-, middle-, and low-frequency information of ${{\bf{I}}_f}$, respectively, to obtain $\{ {\bf{I}}_{f,\xi }^\tau |\tau  \in \{ e,s\} ,\xi  \in \{ h,m,l\} \} $:
\begin{equation}
	\begin{aligned}
		{\bf{I}}_{f,\xi }^e = {{\bf{I}}_f} + \varphi {{\bf{I}}_\xi },
		\qquad
		{\bf{I}}_{f,\xi }^s = {{\bf{I}}_f} - \varphi {{\bf{I}}_\xi },
	\end{aligned}
\end{equation}
where $\varphi $ denotes the adjustment strength, $\xi  \in \{ h,m,l\} $ indicates that the \textbf{h}igh-, \textbf{m}iddle-, or \textbf{l}ow-frequency information in the image is adjusted, and $\tau  \in \{ e,s\} $ indicates that the information of a certain frequency in the image is \textbf{e}nhanced or \textbf{s}uppressed. We feed ${{\bf{I}}_f}$ and ${\bf{I}}_{f,\xi }^\tau $ into the downstream task network, respectively, and obtain the semantic feature ${{\bf{F}}^t}$ and $\{ {\bf{F}}_\xi ^{t,\tau }|\tau  \in \{ e,s\} ,\xi  \in \{ h,m,l\} \} $ fed back by the $t$-th task network. To understand the response degree of the downstream task network to different frequency information in the fused image, we construct the response matrix $\{ d_\xi ^{t,\tau }|\tau  \in \{ e,s\} ,\xi  \in \{ h,m,l\} \} $ using ${{\bf{F}}^t}$ and ${\bf{F}}_\xi ^{t,\tau }$:
\begin{equation}
	\begin{aligned}
		d_\xi ^{t,\tau } = \frac{{{{\left\| {{\bf{F}}_\xi ^{t,\tau } - {{\bf{F}}^t}} \right\|}_2}}}{{{{\left\| {{{\bf{F}}^t}} \right\|}_2} + \varepsilon }},
	\end{aligned}
\end{equation}
where ${\left\|  \cdot  \right\|_2}$ denotes the ${l_2}$-norm, and $\varepsilon $ is a small constant used to avoid a zero denominator. $d_\xi ^{t,\tau }$ denotes the response value. A larger response value indicates that the downstream task network is more sensitive to the information of the $\xi $-frequency, meaning that enhancing or suppressing the information of the $\xi $-frequency will lead to larger performance variations. Based on this idea, we can filter out the sensitive frequency ${\xi ^{t,*}}$ of a specific task:
\begin{equation}
	\begin{aligned}
		{\xi ^{t,*}} = \{ \xi |\mathop {\max }\limits_\xi  (d_\xi ^{t,e},d_\xi ^{t,s}) > \eta \mathop {\max }\limits_{\tau ,\xi } (d_\xi ^{t,\tau })\},
	\end{aligned}
\end{equation}
where $\eta$ is used to control the threshold. However, a larger performance variation does not necessarily indicate a positive performance improvement. Therefore, we further evaluate the information quality of the features ${\bf{F}}_{{\xi ^{t,*}}}^{t,\tau }$. If ${\bf{F}}_{{\xi ^{t,*}}}^{t,e}$ has higher information quality than ${\bf{F}}_{{\xi ^{t,*}}}^{t,s}$ , it indicates that enhancing the sensitive frequency ${\xi ^{t,*}}$ can improve the quality of semantic feature information in the downstream task network and enhance the discriminability of the features, thereby improving downstream task performance. Conversely, it indicates that suppressing the frequency ${\xi ^{t,*}}$ can improve task performance.

To characterize the effective adjustment behavior that can promote task performance improvement as accurately as possible, we introduce three objective evaluation metrics, namely Hoyer Sparsity ${Q_{hs}}$, Negative Entropy ${Q_{ne}}$, and Peak-to-Mean Ratio ${Q_{pr}}$, to evaluate the information quality of ${\bf{F}}_{{\xi ^{t,*}}}^{t,\tau }$ from the perspectives of sparsity, uncertainty, and saliency, respectively. Specifically, a larger ${Q_{hs}}$ indicates that the features are more sparse and contain less redundant information. A larger ${Q_{ne}}$ indicates lower uncertainty in the feature distribution and more concentrated information. A larger ${Q_{pr}}$ indicates stronger feature saliency, suggesting that the information is more prominent. In summary, when these three metrics are simultaneously larger, ${\bf{F}}_{{\xi ^{t,*}}}^{t,\tau }$ usually has stronger discriminability and is more likely to bring positive performance improvement to the downstream task. Based on the above considerations, we define the information quality value $Q_{{\xi ^{t,*}}}^{t,\tau }$ as:
\begin{equation}
	\begin{aligned}
		Q_{{\xi ^{t,*}}}^{t,\tau } = {\tilde Q_{hs}}({\bf{F}}_{{\xi ^{t,*}}}^{t,\tau }) + {\tilde Q_{ne}}({\bf{F}}_{{\xi ^{t,*}}}^{t,\tau }) + {\tilde Q_{pr}}({\bf{F}}_{{\xi ^{t,*}}}^{t,\tau }),
	\end{aligned}
\end{equation}
where ${\tilde Q_{hs}}$, ${\tilde Q_{ne}}$ , and ${\tilde Q_{pr}}$ denote the normalized ${Q_{hs}}$, ${Q_{ne}}$, and ${Q_{pr}}$, respectively. According to $Q_{{\xi ^{t,*}}}^{t,\tau }$, we can select the effective adjustment behavior ${\tau _{{\xi ^{t,*}}}}$ that promotes task performance improvement:
\begin{equation}
	\tau_{\xi^{t,*}} =
	\left\{
	\begin{array}{ll}
		e, & \mathrm{if}\ Q_{\xi^{t,*}}^{t,e} > Q_{\xi^{t,*}}^{t,s}, \\
		s, & \mathrm{otherwise}.
	\end{array}
	\right.
\end{equation}
Based on the above selection process, we can analyze the requirements of arbitrary downstream tasks for different frequency information in the fused image and accordingly construct the causal requirement map  ${{\bf{R}}^t}$. Specifically, ${{\bf{R}}^t}$ contains six channels, each corresponding to a “frequency-adjustment behavior” combination, namely the suppression or enhancement operation for low-, middle-, and high-frequency information. For the effective adjustment behavior ${\tau _{{\xi ^{t,*}}}}$ of the sensitive frequency ${\xi ^{t,*}}$, we multiply the adjusted image ${\bf{I}}_{f,{\xi ^{t,*}}}^{{\tau _{{\xi ^{t,*}}}}}$ by its response value $d_{{\xi ^{t,*}}}^{t,{\tau _{{\xi ^{t,*}}}}}$, and fill the result into the corresponding channel of ${{\bf{R}}^t}$ . For insensitive frequencies or ineffective adjustment behaviors, the corresponding channels in ${{\bf{R}}^t}$ are set to all-zero matrices. In this way, ${\rm{Agent}}_{ir/vi}^v$ can be explicitly informed of the effective frequency, effective adjustment behavior, and response strength required by the current downstream task. ${{\bf{R}}^t}$ is defined as follows:
\begin{equation}
	{\bf{R}}_i^t =
	\left\{
	\begin{array}{ll}
		d_{\xi_i}^{t,\tau_i}{\bf{I}}_{f,\xi_i}^{\tau_i},
		& \mathrm{if}\ \xi_i \in \xi^{t,*},\ \tau_i \in \tau_{\xi^{t,*}}, \\
		\bf{0},
		& \mathrm{otherwise},
	\end{array}
	\right.
	\quad i=1,\ldots,6 .
\end{equation}
where $({\tau _i},{\xi _i})$ denotes the $i$-th “frequency-adjustment behavior” combination in $\{ (e,l),(s,l),(e,m),(s,m),(e,h),(s,h)\} $, and $\bf{0}$ denotes an all-zero matrix. It is worth noting that this paper actively applies different “frequency-adjustment behaviors” to the fused image to observe and interpret the requirements of the task network for high-frequency information, such as fine-grained textures, middle-frequency information, such as object boundaries, and low-frequency information, such as background brightness, in the fused image, thereby constructing a causal requirement map for task-specific requirements. This explicit causal requirement tracing process follows a mechanism of “active intervention $ \to $ response observation $ \to $ requirement interpretation”, which transforms the high-level semantic preferences implicit in the task network into image-level requirement representations, thus achieving the transition from the task-oriented modeling paradigm adopted by existing methods to a requirement-oriented modeling paradigm. Therefore, when facing either seen or unseen tasks, the ${\rm{Agent}}_{ir/vi}^v$ can perceive the differentiated image-level requirements of different tasks.

After obtaining ${{\bf{R}}^t}$, we concatenate it with ${{\bf{I}}_{ir/vi}}$ and $\nabla {{\bf{I}}_{ir/vi}}$. The concatenated features are then used as the state input to ${\rm{Agent}}_{ir/vi}^v$ to obtain the aggregation weight ${\bf{A}}_{ir/vi}^t \in {^{1 \times K}}$:
 \begin{equation}
 	\begin{aligned}
 		{\bf{A}}_{ir/vi}^t = {\rm{Softmax}}({\rm{Agent}}_{ir/vi}^v([{{\bf{R}}^t},{{\bf{I}}_{ir/vi}},\nabla {{\bf{I}}_{ir/vi}}])),
 	\end{aligned}
 \end{equation}
where ${\rm{Agent}}_{ir/vi}^v$ consists of two convolutional layers, one global average pooling layer, and one linear mapping layer. After obtaining ${\bf{A}}_{ir/vi}^t$, we aggregate the requirement knowledge vector $\tilde v_{ir/vi}^t \in {^{1 \times e}}$, which serves the specific requirement of the downstream task, from the ${{\bf{V}}_{ir/vi}}$ according to ${\bf{A}}_{ir/vi}^t$:
 \begin{equation}
	\begin{aligned}
		\tilde v_{ir/vi}^t = {\bf{A}}_{ir/vi}^t{{\bf{V}}_{ir/vi}}.
	\end{aligned}
\end{equation}
To enable ${\rm{Agent}}_{ir/vi}^v$ to learn the ability to accurately model task requirements, so that $\tilde v_{ir/vi}^t$ can more precisely represent task-specific requirements and thus improve downstream task performance after semantic restoration, we provide a reward ${r^v}$ to ${\rm{Agent}}_{ir/vi}^v$ according to the performance improvement of the downstream task before and after semantic restoration:
 \begin{equation}
	\begin{aligned}
		{r^v} = \sum\limits_{t = 1}^N {\max (0,{c^t}({\bf{\hat y}}_o^t,{\bf{y}}_{gt}^t) - {c^t}({{{\bf{\hat y}}}^t},{\bf{y}}_{gt}^t))},
	\end{aligned}
\end{equation}
where ${\bf{\hat y}}_o^t$ denotes the task result of the $t$-th downstream task before semantic restoration. If the downstream task loss after semantic restoration is smaller than that before semantic restoration, it indicates that $\tilde v_{ir/vi}^t$ obtained by ${\rm{Agent}}_{ir/vi}^v$ can accurately describe the task requirement, enabling semantic restoration to positively contribute to the improvement of task performance. In this case, ${r^v} > 0$, indicating that a certain reward is assigned to ${\rm{Agent}}_{ir/vi}^v$, and the reward increases with the improvement of task performance. Otherwise, ${r^v} = 0$, and no reward is assigned to ${\rm{Agent}}_{ir/vi}^v$.

\subsection{History-Analysis Multi-Objective Balancing}\label{sec3_3}

The History-Analysis Multi-Objective Balancing mechanism introduces a task balancing agent (${\rm{Agent}^l}$) to monitor multi-task performance improvements and imbalance states, thereby adaptively regulating ${\ell _{ta}}$ to ensure multi-task balance. Specifically, we record the multi-task performance improvement magnitudes and their variance over the most recent $X$ steps during the historical optimization process, which are collectively referred to as historical performance statistics. Among them, the multi-task performance improvement magnitude provides two types of information, namely the performance improvement magnitude $\Delta P_{io,x}^t$ relative to that before semantic restoration and the performance improvement magnitude $\Delta P_{ip,x}^t$ relative to that after semantic restoration in the previous step:
 \begin{equation}
	\begin{aligned}
		\Delta P_{io,x}^t = {{(\ell _o^t - \ell _x^t)} \mathord{\left/
				{\vphantom {{(\ell _o^t - \ell _x^t)} {\ell _o^t}}} \right.
				\kern-\nulldelimiterspace} {\ell _o^t}},
		\qquad
		\Delta P_{ip,x}^t = {{(\ell _{x - 1}^t - \ell _x^t)} \mathord{\left/
				{\vphantom {{(\ell _{x - 1}^t - \ell _x^t)} {\ell _{x - 1}^t}}} \right.
				\kern-\nulldelimiterspace} {\ell _{x - 1}^t}},		
	\end{aligned}
\end{equation}
where $\ell _x^t$ denotes the loss value of the $t$-th task on the fused image after semantic restoration in the previous $x$-th step, where $x(x = 1,2, \ldots ,X)$, and $\ell _x^t = {c^t}({\bf{\hat y}}_x^t,{\bf{y}}_{gt}^t)$. ${\bf{\hat y}}_x^t$ denotes the task result of the fused image after semantic restoration in the previous $x$-th step. $\ell _o^t$ denotes the loss value of the $t$-th task on the fused image before semantic restoration, and $\ell _o^t = {c^t}({\bf{\hat y}}_o^t,{\bf{y}}_{gt}^t)$. To more intuitively reflect the imbalance in the multi-task optimization process, we calculate the variance $\sigma _{io/ip,x}^2$ of $\Delta P_{io/ip,x}^t$:
 \begin{equation}
	\begin{aligned}
		\sigma _{io/ip,x}^2 = \frac{1}{N}\sum\limits_{t = 1}^N {{{(\Delta P_{io/ip,x}^t - \Delta \bar P_{io/ip,x}^{})}^2}},
	\end{aligned}
\end{equation}
 \begin{equation}
	\begin{aligned}
		\Delta \bar P_{io/ip,x}^{} = \frac{1}{N}\sum\limits_{t = 1}^N {\Delta P_{io/ip,x}^t}.
	\end{aligned}
\end{equation}
To alleviate the imbalance caused by multi-task optimization, we take $\Delta P_{io/ip,x}^t$ and $\sigma _{io/ip,x}^2$ as the current state and feed them into ${\rm{Agent}^l}$, enabling it to monitor the multi-task optimization process and generate the balancing parameters $\beta  = \{ {\beta ^t}\} _{t = 1}^N$ for dynamically regulating the ${\ell _{ta}}$:
 \begin{equation}
 	\resizebox{1\columnwidth}{!}{$
 	\displaystyle
	\begin{aligned}
\beta  = {\rm{Agen}}{{\rm{t}}^l}([&\{ \Delta P_{ip,x}^1\} _{x = 1}^X,\{ \Delta P_{io,x}^1\} _{x = 1}^X, \ldots ,\{ \Delta P_{ip,x}^N\} _{x = 1}^X,\{ \Delta P_{io,x}^N\} _{x = 1}^X,\\
&\{ \sigma _{ip,x}^2\} _{x = 1}^X,\{ \sigma _{io,x}^2\} _{x = 1}^X]),
	\end{aligned}
	$}
\end{equation}
where ${\rm{Agent}^l}$ consists of two linear mapping layers.

\begin{figure}
	\begin{center}
		\includegraphics[width=0.96\linewidth]{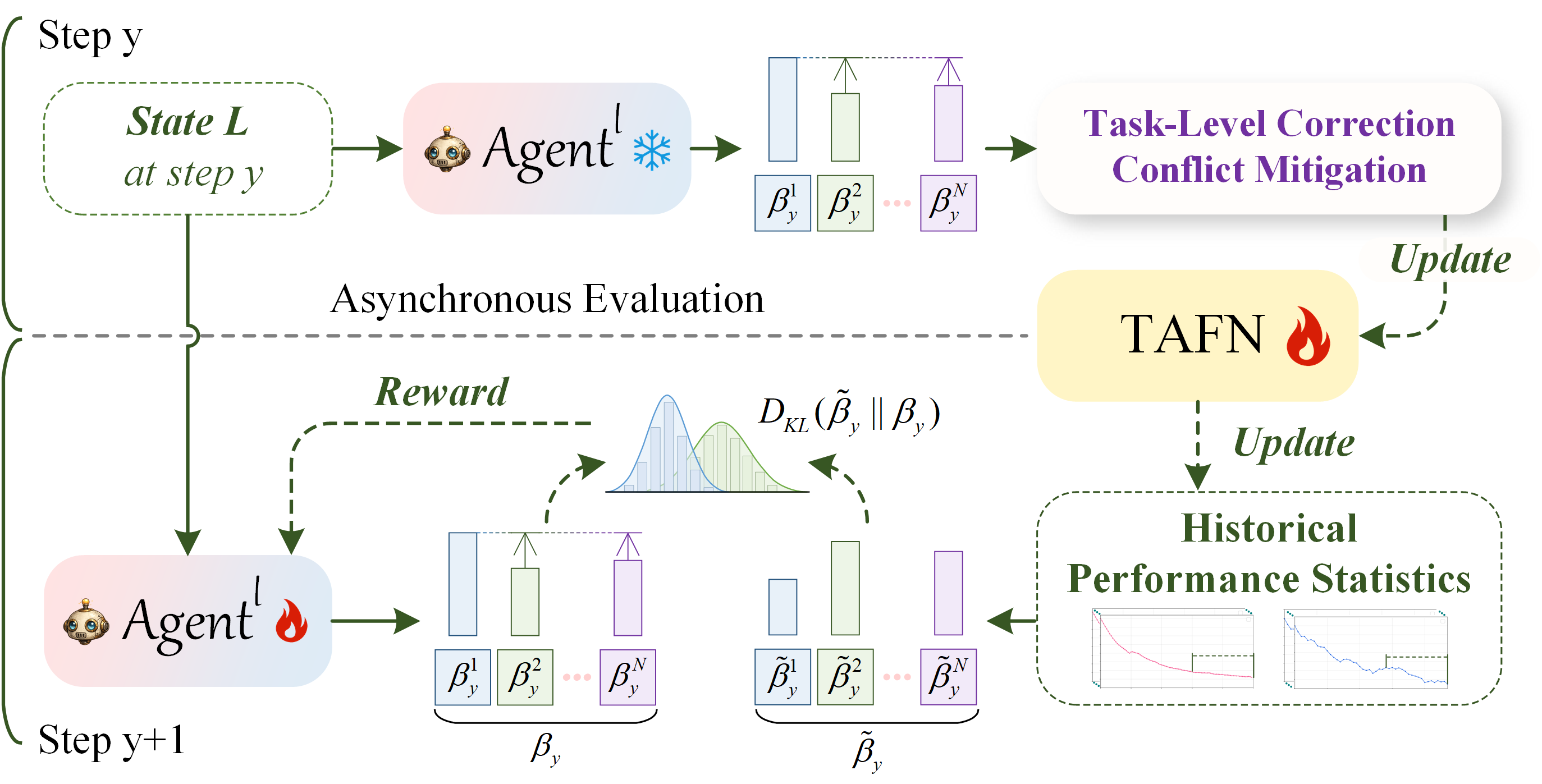}
	\end{center}
	\caption{
		Asynchronous evaluation strategy of ${\rm{Agent}^l}$.
	}\vspace{-2mm}
	\label{fig5}
\end{figure}

However, the evaluation of multi-task balance often requires statistical data based on long-term multi-task performance variations, making it difficult to evaluate the behavior of ${\rm{Agent}^l}$. Therefore, we design an asynchronous evaluation strategy to evaluate the behavior of ${\rm{Agent}^l}$. As shown in Fig. \ref{fig5}, we feed the state at the $y$-th step into ${\rm{Agent}^l}$, and input the generated ${\beta _y}$ at the $y$-th step into the subsequent Task-Level-Correction Conflict Mitigation mechanism to update TAFN and the historical performance statistics. Based on the updated historical performance statistics, we calculate the pseudo ground truth ${\tilde \beta _y} = \{ \tilde \beta _y^t\} _{t = 1}^N$ at the $y$-th step:
 \begin{equation}
	\begin{aligned}
		{\tilde \beta _y} = \mathop {{\rm{Softmax}}}\limits_t {\rm{( - }}\frac{1}{M}\sum\limits_{x = y - M}^y {\Delta P_{ip,x}^t} {\rm{)}},
	\end{aligned}
\end{equation}
where ${1 \mathord{\left/
		{\vphantom {1 M}} \right.
		\kern-\nulldelimiterspace} M}\sum\nolimits_{x = y - M}^y {\Delta P_{ip,x}^t} $ denotes the average performance improvement of the $t$-th task over the previous $M$ steps. A smaller value indicates that the performance improvement of the $t$-th task is smaller and thus requires more emphasis and attention. Therefore, the ${\tilde \beta _y}$ can reflect multi-task imbalance to a certain extent. We compute the negative KL divergence between ${\beta _y}$ and ${\tilde \beta _y}$ as the reward ${r^l}$ for ${\rm{Agent}^l}$:
 \begin{equation}
	\begin{aligned}
		{r^l} =  - {D_{KL}}({\tilde \beta _y}||{\beta _y}) =  - \sum\limits_{t = 1}^N {\tilde \beta _y^t(\log \tilde \beta _y^t - \log \beta _y^t)}.
	\end{aligned}
\end{equation}
In this way, after receiving the performance feedback, the action taken by ${\rm{Agent}^l}$ at the $y$-th step is evaluated and optimized asynchronously by constructing the pseudo ground truth at the $y+1$-th step. The optimized ${\rm{Agent}^l}$ is then used to monitor the imbalance state at the $y+2$-th step.

\subsection{Task-Level-Correction Conflict Mitigation}\label{sec3_4}
The Task-Level-Correction Conflict Mitigation mechanism employs a conflict mitigation agent (${\rm{Agent}^g}$) to analyze the conflict and synergy relationships between task-specific gradients and the multi-task joint gradient, thereby correcting the multi-task joint gradient with task-specific gradients to alleviate gradient conflicts. Specifically, we compute the gradients of TAFN using the multi-task joint task-aware loss ${\ell _{ta}}$ and the loss ${\ell ^t}$ of each specific task, respectively, thereby obtaining the multi-task joint gradient ${{\bf{g}}_a}$ and the task-specific gradients $\{ {{\bf{g}}^t}\} _{t = 1}^N$. We then calculate the multi-task gradient relationships among them. The multi-task gradient relationships consist of two parts, namely the inter-task gradient relationships ${s_{t1,t2}}$ and the relationships between the multi-task joint gradient and each task-specific gradient ${s_{a,t}}$:
\begin{equation}
	\operatorname{sim}(\bf{u},\bf{v})
	=
	\frac{\langle \bf{u},\bf{v}\rangle}
	{\|\bf{u}\|_2\|\bf{v}\|_2},
\end{equation}
\begin{equation}
	s_{t1,t2}
	=
	\operatorname{sim}({\bf{g}}^{t1},{\bf{g}}^{t2}),
	\qquad
	s_{a,t}
	=
	\operatorname{sim}({\bf{g}}_{a},{\bf{g}}^{t}).
\end{equation}
where $\left\langle  \cdot  \right\rangle $ denotes the inner product of vectors, and ${s_{t1,t2}}$ denotes the cosine similarity between the gradients of the $t1$-th task and the $t2$-th task. When ${s_{t1,t2}} > 0$, the two gradients have consistent directions and exhibit a synergistic relationship. When ${s_{t1,t2}} < 0$, the two gradients have opposite directions and exhibit a conflicting relationship. When ${s_{t1,t2}} = 0$, the two gradients are approximately orthogonal and exhibit no explicit conflict or synergy. To enable ${\rm{Agent}^g}$ to more comprehensively model the gradient conflicts and synergies among multiple tasks as well as the relationships among multi-task performance, we take the multi-task gradient relationships, gradient norms, and performance improvements of all tasks as the state input to ${\rm{Agent}^g}$, thereby obtaining the gradient correction weight $\chi $:
 \begin{equation}
	\begin{aligned}
		\chi  = {\rm{Agen}}{{\rm{t}}^g}([&\{ {s_{a,t}}\} _{t = 1}^N,\{ {s_{t1,t2}}\} _{t1 = 1,t2 = 1}^{N,N}, \\
		&||{{\bf{g}}_a}|{|_2},\{ ||{{\bf{g}}^t}|{|_2}\} _{t = 1}^N,\\
		&\{ \Delta P_{io,1}^t\} _{t = 1}^N,\{ \Delta P_{ip,1}^t\} _{t = 1}^N]).
	\end{aligned}
\end{equation}
After obtaining $\chi $, we decompose the task-specific gradients $\{ {{\bf{g}}^t}\} _{t = 1}^N$ into vertical components $\{ {\bf{g}}_v^t\} _{t = 1}^N$ and horizontal components $\{ {\bf{g}}_h^t\} _{t = 1}^N$, and aggregate them using $\chi $ to correct ${{\bf{g}}_a}$. Finally, we obtain the corrected gradient ${{\bf{g}}_u}$:
 \begin{equation}
	\begin{aligned}
		{{\bf{g}}_u} = {{\bf{g}}_a} + \sum\limits_{t = 1}^N {(\chi _v^t{\bf{g}}_v^t + \chi _h^t{\bf{g}}_h^t)}.
	\end{aligned}
\end{equation}
Finally, we update the parameters of TAFN using the ${{\bf{g}}_u}$. To encourage $\mathrm{Agent}^{g}$ to effectively correct ${\bf{g}}_{a}$, we design the reward $r^{g}$ as follows:
 \begin{equation}
	\begin{aligned}
		{r^g} = \frac{1}{N}\sum\limits_{t = 1}^N {{\operatorname{sim}({\bf{g}}_{u},{\bf{g}}^{t})}}  + \mathop {\min }\limits_t ({\operatorname{sim}({\bf{g}}_{u},{\bf{g}}^{t})}),
	\end{aligned}
\end{equation}
where this formula includes the mean and minimum cosine similarities between the corrected gradient ${{\bf{g}}_u}$ and the task-specific gradients $\{ {{\bf{g}}^t}\} _{t = 1}^N$. Larger values indicate smaller gradient conflicts in the multi-task optimization process, and thus ${\rm{Agent}^g}$ receives a larger reward ${r^g}$.

\begin{figure*}
	\begin{center}
		\includegraphics[width=0.96\linewidth]{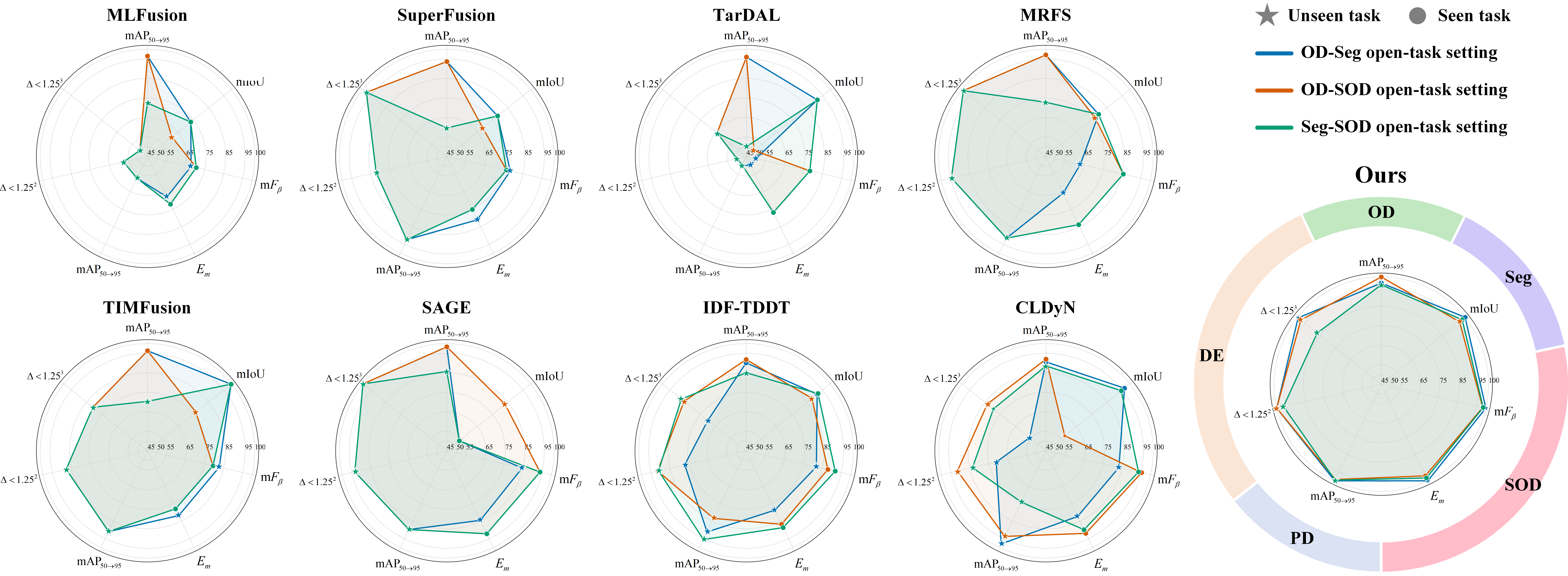}
	\end{center}
	\caption{Quantitative visualization results of the proposed method compared with task-network retraining methods and multi-task-aware methods in diverse open-task scenarios.}\vspace{-2mm}
	\label{fig6}
\end{figure*}

\section{Experiments}\label{sec4}
\subsection{Datasets}
To verify the downstream task adaptability of the proposed method in open-task scenarios, we train the model on two downstream task datasets and evaluate its task generalization performance on three unseen downstream task datasets. Specifically, according to the open-task scenario setting, two datasets are selected from M$^3$FD \cite{8}, FMB \cite{6}, and VT5000 \cite{109} for training, and the remaining dataset that is not involved in training, together with LLVIP \cite{50} and VTD \cite{133}, is used for testing. Among them, M$^3$FD, FMB, VT5000, LLVIP, and VTD correspond to \textbf{object detection (OD)}, \textbf{semantic segmentation (Seg)}, \textbf{salient object detection (SOD)}, \textbf{pedestrian detection (PD)}, and \textbf{depth estimation (DE)}, respectively. For datasets with official data splits, we adopt their official training and test sets, including FMB (train: 1,220 pairs, test: 280 pairs), VT5000 (train: 2,500 pairs, test: 2,500 pairs), LLVIP (test: 3,463 pairs), and VTD (test: 1,928 pairs). For datasets without official splits, we follow the commonly used setting in the field and randomly split the data, such as M$^3$FD (train: 3,150 pairs, test: 525 pairs). In addition, according to the task categories corresponding to the selected training sets, we augment the training samples by referring to the data augmentation strategies used in YOLOv5\footnote{https://github.com/ultralytics/yolov5} for OD, SegFormer \cite{110} for Seg, and CTDNet \cite{111} for SOD, respectively.

\subsection{Implementation Details}
During training, we use the AdamW optimizer for parameter optimization. The batch size is set to 3, the learning rate is set to ${10^{ - 2}}$, and the total number of training epochs is 160. The main hyperparameters $K$, $e$, $\varphi $, and $\eta $ are set to 128, 128, 0.2, and 0.7, respectively. During training and validation, we adopt the widely used YOLOv5s, SegFormer (mit-b2), CTDNet-18, and Lite-Mono$_{Base}$ \cite{134} as downstream task networks. The proposed method is implemented based on the PyTorch framework and trained on a single NVIDIA GeForce RTX 4090 GPU.

\subsection{Evaluation Metrics}
In the experiments, we select ten commonly used objective evaluation metrics to comprehensively evaluate the proposed method from two aspects: performance in open-task scenarios and fusion performance. For downstream task performance evaluation, we use mean Average Precision at intersection-over-union (IoU) thresholds from 0.5 to 0.95 (${\rm{mA}}{{\rm{P}}_{50 \to 95}}$) \cite{127} to evaluate the performance of OD and PD, mean pixel Intersection over Union (mIoU) \cite{114} to evaluate Seg performance, mean F-measure (${\rm{m}}{F_\beta }$) \cite{115} and E-measure (${E_m}$) \cite{116} to evaluate SOD performance, and depth ratio accuracy ($\Delta  < {1.25^2}$ and $\Delta  < {1.25^3}$) \cite{137} to evaluate DE performance. For fusion performance evaluation, we adopt the Chen-Blum Metric (${Q_{CB}}$) \cite{135}, Structural Similarity Index Measure (SSIM) \cite{48}, Correlation Coefficient (CC) \cite{55}, and Fusion Artifact Metric (${N_{AB/F}}$) \cite{136} to evaluate the quality of fused images from the perspectives of human visual perception quality, structural similarity, correlation with source images, and fusion artifacts, respectively. It should be noted that $N_{AB/F}$ is a negative metric, where smaller values indicate better performance, whereas all the other evaluation metrics are positive metrics, where larger values indicate better performance.

\begin{table}[htbp]
	\centering
	\caption{Quantitative comparison of the proposed method with task-network retraining methods and multi-task-aware methods in the \textbf{OD-Seg} open-task scenario. The best, second-best, and third-best metric values are highlighted with {\setlength{\fboxsep}{2.3pt}\colorbox{lightred}{Red}}, {\setlength{\fboxsep}{2.3pt}\colorbox{lightblue}{Blue}}, and {\setlength{\fboxsep}{2.3pt}\colorbox{lightgreen}{Green}} backgrounds, respectively.}
	\renewcommand{\arraystretch}{1.6}
	\setlength{\tabcolsep}{2pt}
	\scriptsize 
	\begin{tabular}{c|c|c|cc|c|cc}
		\toprule
		\multirow{3}{*}{\textbf{Methods}} & \multicolumn{2}{c|}{\textbf{Seen Tasks}} & \multicolumn{5}{c}{\textbf{Unseen Tasks}} \\
		\cline{2-8}          & \textbf{OD}   & \textbf{Seg}   & \multicolumn{2}{c|}{\textbf{SOD}} & \textbf{PD}    & \multicolumn{2}{c}{\textbf{DE}} \\
		\cline{2-8}          & ${\rm{mA}}{{\rm{P}}_{50 \to 95}}$ & mIoU  & ${\rm{m}}{F_\beta }$    & ${E_m}$    & ${\rm{mA}}{{\rm{P}}_{50 \to 95}}$ & $\Delta  < {1.25^2}$    & $\Delta  < {1.25^3}$ \\
		\midrule
		MLFusion & 0.6172  & 57.47  & 0.7900  & 0.8950  & 0.4703  & 0.8681  & 0.9423  \\
		SuperFusion & 0.6089  & 58.11  & 0.7977  & 0.9013  & \cellcolor{lightgreen}{0.6019} & 0.8777  & 0.9530  \\
		TarDAL & 0.6157  & 59.80  & 0.7767  & 0.8864  & 0.4424  & 0.8651  & 0.9455  \\
		MRFS  & \cellcolor{lightblue}{0.6189} & 58.28  & 0.7865  & 0.8939  & 0.5985  & \cellcolor{lightblue}{0.8826} & \cellcolor{lightgreen}{0.9533} \\
		TIMFusion & 0.6166  & \cellcolor{lightblue}{60.86} & 0.8009  & 0.9017  & 0.5965  & 0.8799  & 0.9492  \\
		SAGE  & \cellcolor{lightred}{0.6225} & 54.89  & \cellcolor{lightblue}{0.8024} & \cellcolor{lightblue}{0.9030} & 0.5921  & \cellcolor{lightgreen}{0.8821} & \cellcolor{lightred}{0.9535} \\
		\hline
		IDF-TDDT & 0.5985  & 59.81  & 0.8004  & 0.9002  & 0.5973  & 0.8758  & 0.9468  \\
		CLDyN & 0.5999  & \cellcolor{lightgreen}{60.44} & \cellcolor{lightgreen}{0.8014} & \cellcolor{lightgreen}{0.9019} & \cellcolor{lightblue}{0.6229} & 0.8734  & 0.9436  \\
		\hline
		Ours  & \cellcolor{lightgreen}{0.6184} & \cellcolor{lightred}{60.89} & \cellcolor{lightred}{0.8137} & \cellcolor{lightred}{0.9103} & \cellcolor{lightred}{0.6304} & \cellcolor{lightred}{0.8847} & \cellcolor{lightblue}{0.9534} \\
		\bottomrule
	\end{tabular}
	\label{tab1}%
\end{table}%

\subsection{Open-Task Adaptability Comparison}

To verify the task generalization ability of the proposed method in open-task scenarios, we compare it with two representative categories of task-aware methods, including task-network retraining methods and multi-task-aware methods. Specifically, task-network retraining methods retrain task networks based on fused images to adapt them to specific task requirements. If such methods are required to adapt to multiple tasks, multiple retraining processes are needed. In contrast, multi-task-aware methods optimize the fusion network using joint constraints from multiple tasks, thereby improving its adaptability to diverse task requirements. To comprehensively evaluate task generalization ability in open-task scenarios, we construct diverse open-task scenarios. Specifically, two tasks are selected from OD, Seg, and SOD as seen tasks for model training, while the remaining task, together with PD and DE, is used as unseen tasks for performance evaluation. Based on this setting, we construct three open-task scenarios, namely \textbf{“OD + Seg $ \to $ Seen, SOD + PD + DE $ \to $ Unseen”, “OD + SOD $ \to $ Seen, Seg + PD + DE $ \to $ Unseen”, and “Seg + SOD $ \to $ Seen, OD + PD + DE $ \to $ Unseen”, denoted as OD-Seg, OD-SOD, and Seg-SOD}, respectively.

\begin{figure*}
	\begin{center}
		\includegraphics[width=0.96\linewidth]{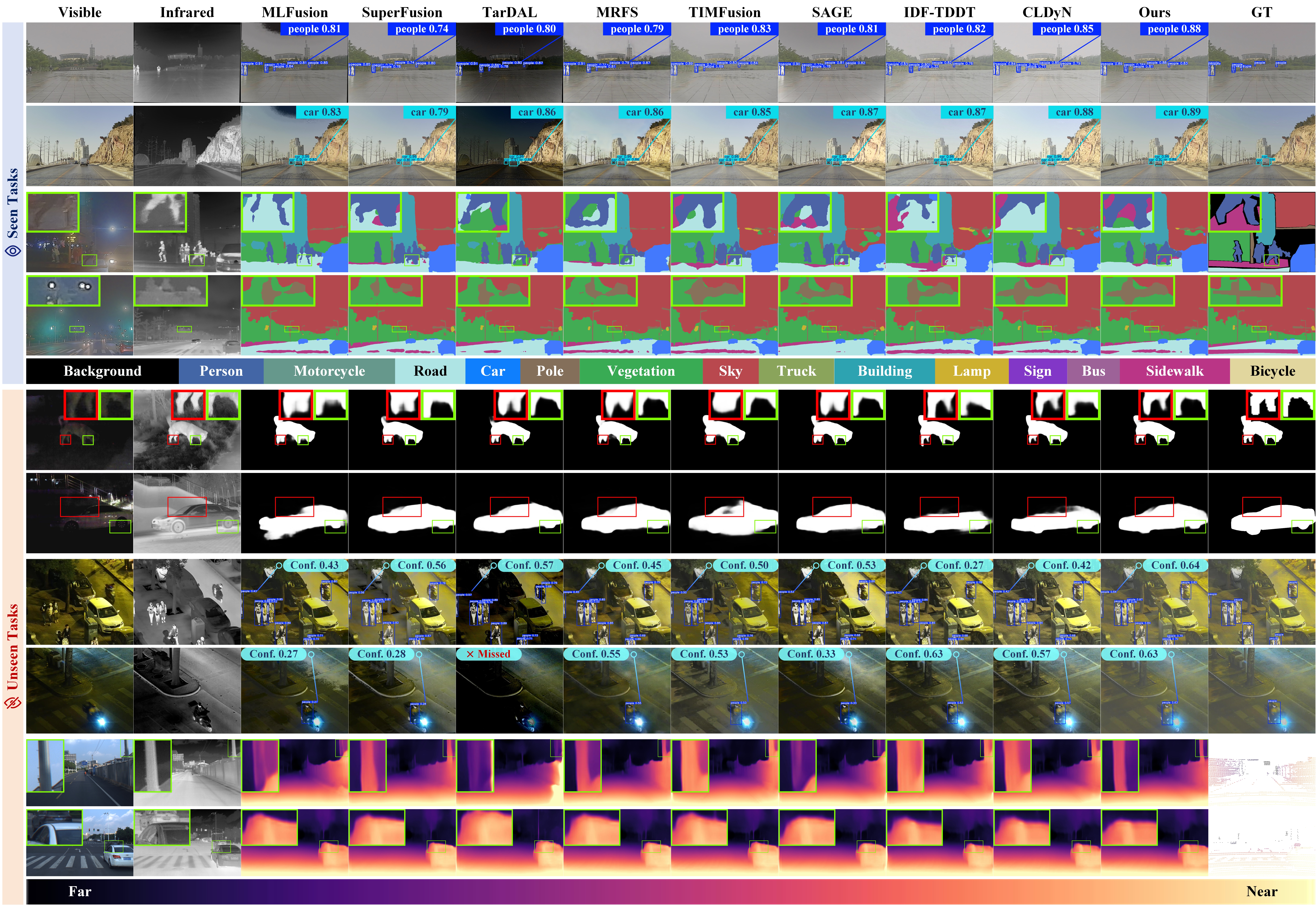}
	\end{center}
	\caption{Qualitative comparison of the proposed method with task-network retraining methods and multi-task-aware methods in the \textbf{OD-Seg} open-task scenario. The first and second columns show the visible and infrared source images, respectively. The third to eleventh columns present the downstream task results of the compared methods, and the twelfth column shows the Ground Truth (GT). Every two rows correspond to one task, showing the results of OD, Seg, SOD, PD, and DE from top to bottom, respectively.}\vspace{-2mm}
	\label{fig7}
\end{figure*}

\begin{figure*}
	\begin{center}
		\includegraphics[width=0.96\linewidth]{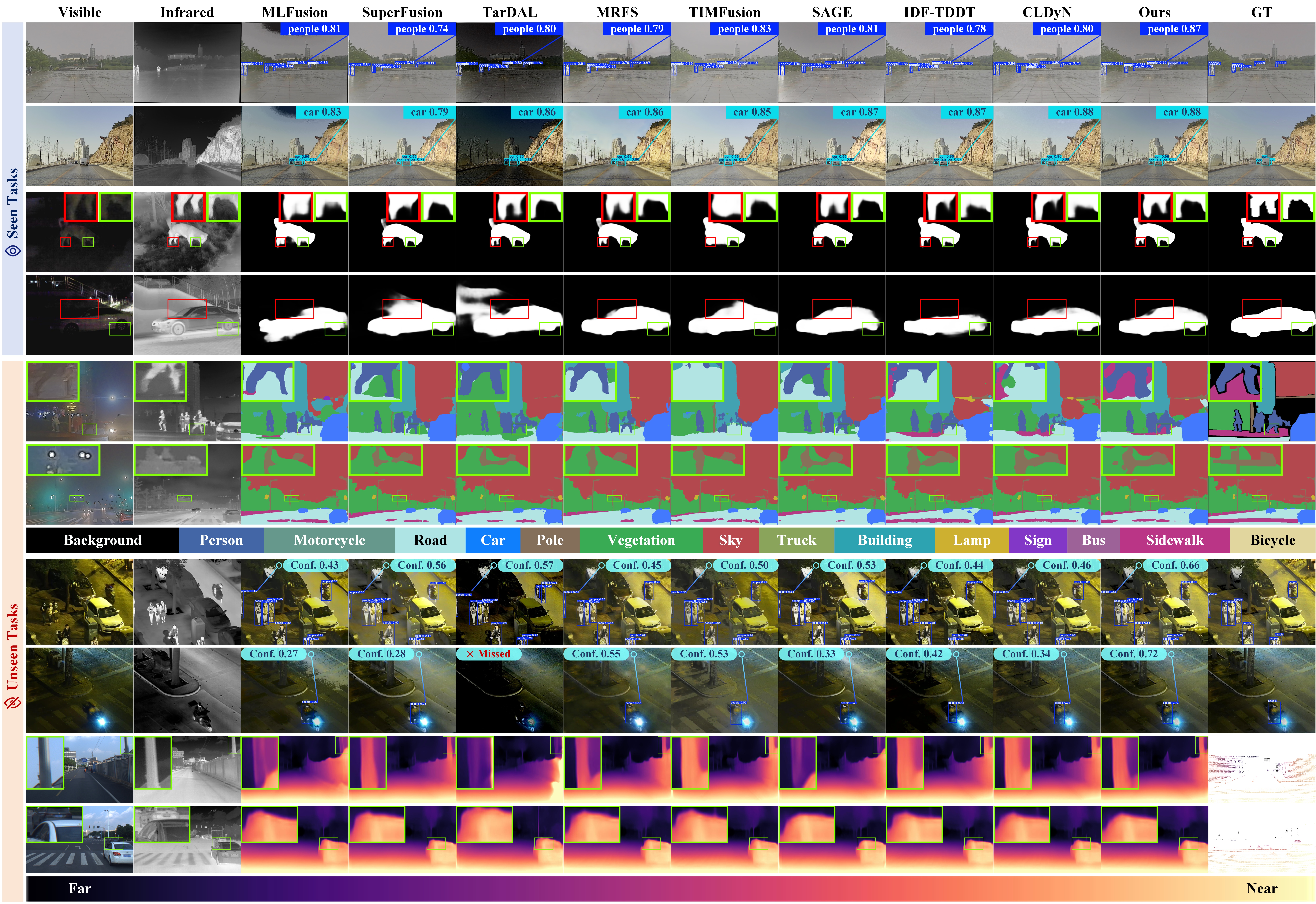}
	\end{center}
	\caption{Qualitative comparison of the proposed method with task-network retraining methods and multi-task-aware methods in the \textbf{OD-SOD} open-task scenario. The first and second columns show the visible and infrared source images, respectively. The third to eleventh columns present the downstream task results of the compared methods, and the twelfth column shows the Ground Truth (GT). Every two rows correspond to one task, showing the results of OD, Seg, SOD, PD, and DE from top to bottom, respectively.}
	\label{fig8}
\end{figure*}

\begin{figure*}
	\begin{center}
		\includegraphics[width=0.96\linewidth]{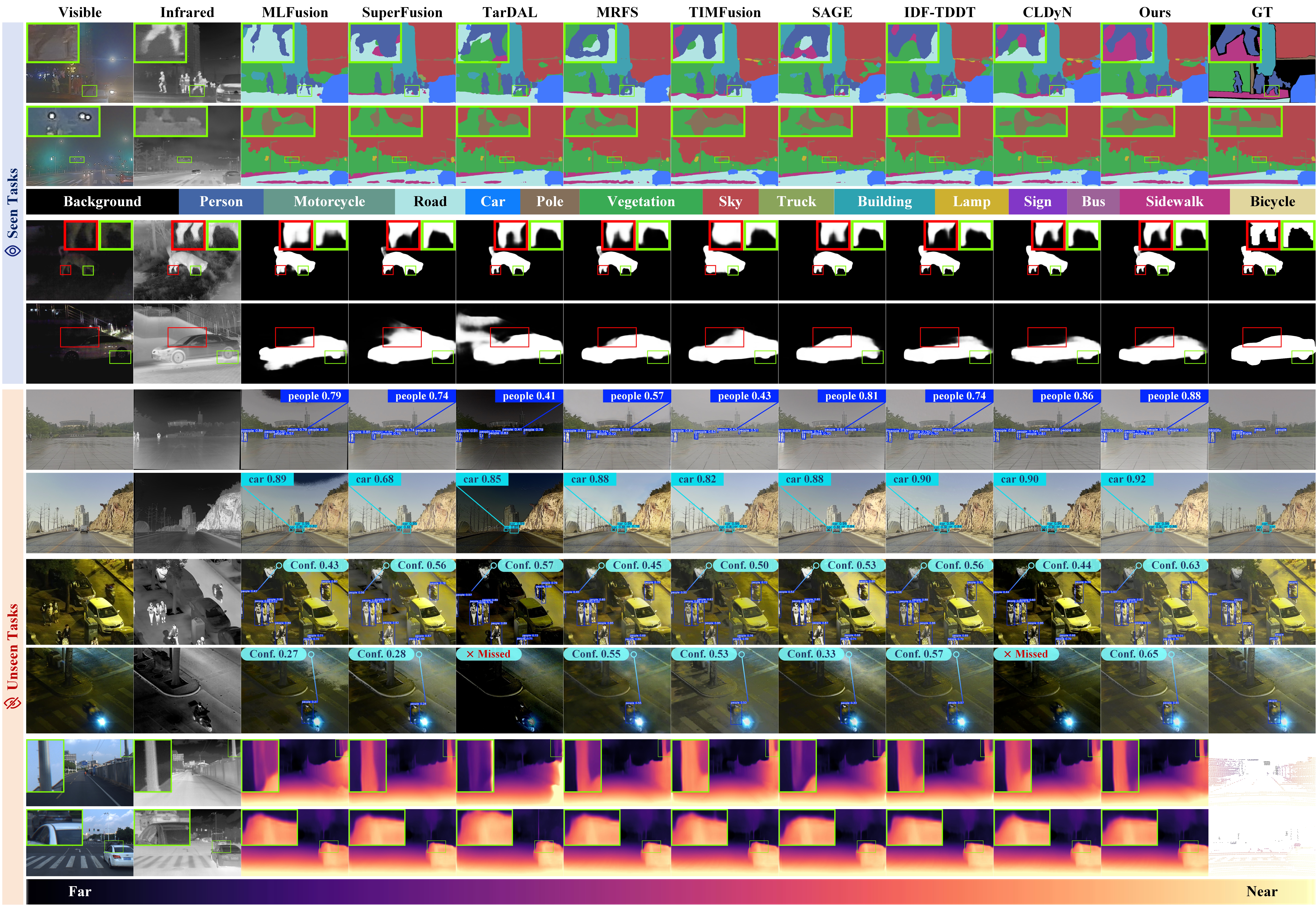}
	\end{center}
	\caption{Qualitative comparison of the proposed method with task-network retraining methods and multi-task-aware methods in the \textbf{Seg-SOD} open-task scenario. The first and second columns show the visible and infrared source images, respectively. The third to eleventh columns present the downstream task results of the compared methods, and the twelfth column shows the Ground Truth (GT). Every two rows correspond to one task, showing the results of OD, Seg, SOD, PD, and DE from top to bottom, respectively.}
	\label{fig9}
\end{figure*}

\begin{table}[htbp]
	\centering
	\caption{Quantitative comparison of the proposed method with task-network retraining methods and multi-task-aware methods in the \textbf{OD-SOD} open-task scenario. The best, second-best, and third-best metric values are highlighted with {\setlength{\fboxsep}{2.3pt}\colorbox{lightred}{Red}}, {\setlength{\fboxsep}{2.3pt}\colorbox{lightblue}{Blue}}, and {\setlength{\fboxsep}{2.3pt}\colorbox{lightgreen}{Green}} backgrounds, respectively.}
	\renewcommand{\arraystretch}{1.6}
	\setlength{\tabcolsep}{1.5pt}
	\scriptsize 
	\begin{tabular}{c|c|cc|c|c|cc}
		\toprule
		\multirow{3}{*}{\textbf{Methods}} & \multicolumn{3}{c|}{\textbf{Seen Tasks}} & \multicolumn{4}{c}{\textbf{Unseen Tasks}} \\
		\cline{2-8}          & \textbf{OD}    & \multicolumn{2}{c|}{\textbf{SOD}} & \textbf{Seg}   & \textbf{PD}    & \multicolumn{2}{c}{\textbf{DE}} \\
		\cline{2-8}          & ${\rm{mA}}{{\rm{P}}_{50 \to 95}}$ & ${\rm{m}}{F_\beta }$    & ${E_m}$    & mIoU  & ${\rm{mA}}{{\rm{P}}_{50 \to 95}}$ & $\Delta  < {1.25^2}$    & $\Delta  < {1.25^3}$ \\
		\midrule
		MLFusion & 0.6172  & 0.7920  & 0.8970  & 55.89  & 0.4703  & 0.8681  & 0.9423  \\
		SuperFusion & 0.6089  & 0.7963  & 0.8984  & 56.86  & \cellcolor{lightgreen}{0.6019} & 0.8777  & 0.9530  \\
		TarDAL & 0.6157  & 0.7977  & 0.8993  & 54.48  & 0.4424  & 0.8651  & 0.9455  \\
		MRFS  & \cellcolor{lightgreen}{0.6189} & 0.8032  & 0.9026  & 57.96  & 0.5985  & \cellcolor{lightblue}{0.8826} & \cellcolor{lightblue}{0.9533} \\
		TIMFusion & 0.6166  & 0.7985  & 0.8998  & 57.91  & 0.5965  & 0.8799  & 0.9492  \\
		SAGE  & \cellcolor{lightblue}{0.6225} & \cellcolor{lightgreen}{0.8093} & \cellcolor{lightblue}{0.9066} & \cellcolor{lightgreen}{58.74} & 0.5921  & \cellcolor{lightgreen}{0.8821} & \cellcolor{lightred}{0.9535} \\
		\hline
		IDF-TDDT & 0.6035  & 0.8048  & 0.9040  & \cellcolor{lightblue}{59.37} & 0.5682  & 0.8814  & 0.9503  \\
		CLDyN & 0.6040  & \cellcolor{lightblue}{0.8103} & \cellcolor{lightgreen}{0.9065} & 55.42  & \cellcolor{lightblue}{0.6068} & 0.8814  & 0.9498  \\
		\hline
		Ours  & \cellcolor{lightred}{0.6274} & \cellcolor{lightred}{0.8125} & \cellcolor{lightred}{0.9090} & \cellcolor{lightred}{60.42} & \cellcolor{lightred}{0.6275} & \cellcolor{lightred}{0.8847} & \cellcolor{lightgreen}{0.9530} \\
		\bottomrule
	\end{tabular}
	\label{tab2}%
\end{table}%

\subsubsection{Comparison with Task-Network Retraining Methods}

We compare the proposed method with task-network retraining methods in diverse open-task scenarios to verify its task generalization ability. Specifically, we use the fused images generated by MLFusion \cite{1}, SuperFusion \cite{11}, TarDAL \cite{8}, MRFS \cite{90}, TIMFusion \cite{65}, and SAGE \cite{73} to retrain the networks of seen tasks, respectively. In the evaluation of seen tasks, the fused images are deployed to the corresponding retrained task networks. In the evaluation of unseen tasks, the fused images are directly deployed to the task networks without retraining. As shown in Tables \ref{tab1}, \ref{tab2}, and \ref{tab3}, under different open-task scenarios, the proposed method achieves performance comparable to or even better than that of task-network retraining methods on multiple seen tasks, while exhibiting better generalization ability on unseen tasks. In addition, the proposed method consistently demonstrates favorable performance regardless of whether the task is seen or unseen. In contrast, task-network retraining methods usually suffer from significant performance degradation on unseen tasks. For example, for MLFusion, when OD is a seen task, reaches 0.6172, whereas when OD is an unseen task, its performance drops to 0.5467, indicating that this type of method struggles to generalize effectively to unseen tasks. As shown in Fig. \ref{fig6}, we further visualize the quantitative analysis results using radar charts, which more intuitively show that the proposed method can achieve favorable performance regardless of whether the task is seen or unseen. Moreover, it demonstrates strong task adaptability across diverse open-task scenarios composed of five downstream tasks.

We further conduct qualitative analysis. As shown in Figs. \ref{fig7}, \ref{fig8}, and \ref{fig9}, the proposed method produces higher-confidence detection results in OD and PD tasks, better preserves fine-grained object structures in Seg and SOD tasks, and effectively alleviates foreground-background depth confusion in the DE task. These results further demonstrate the stronger open-task generalization ability of the proposed method.

\begin{table}[htbp]
	\centering
	\caption{Quantitative comparison of the proposed method with task-network retraining methods and multi-task-aware methods in the \textbf{Seg-SOD} open-task scenario. The best, second-best, and third-best metric values are highlighted with {\setlength{\fboxsep}{2.3pt}\colorbox{lightred}{Red}}, {\setlength{\fboxsep}{2.3pt}\colorbox{lightblue}{Blue}}, and {\setlength{\fboxsep}{2.3pt}\colorbox{lightgreen}{Green}} backgrounds, respectively.}
	\renewcommand{\arraystretch}{1.6}
	\setlength{\tabcolsep}{1.5pt}
	\scriptsize 
	\begin{tabular}{c|c|cc|c|c|cc}
		\toprule
		\multirow{3}{*}{\textbf{Methods}} & \multicolumn{3}{c|}{\textbf{Seen Tasks}} & \multicolumn{4}{c}{\textbf{Unseen Tasks}} \\
		\cline{2-8}          & \textbf{Seg}   & \multicolumn{2}{c|}{\textbf{SOD}} & \textbf{OD}    & \textbf{PD}    & \multicolumn{2}{c}{\textbf{DE}} \\
		\cline{2-8}          & mIoU  & ${\rm{m}}{F_\beta }$    & ${E_m}$    & ${\rm{mA}}{{\rm{P}}_{50 \to 95}}$ & ${\rm{mA}}{{\rm{P}}_{50 \to 95}}$ & $\Delta  < {1.25^2}$    & $\Delta  < {1.25^3}$ \\
		\midrule
		MLFusion & 57.47  & 0.7920  & 0.8970  & 0.5467  & 0.4703  & 0.8681  & 0.9423  \\
		SuperFusion & 58.11  & 0.7963  & 0.8984  & 0.5093  & \cellcolor{lightgreen}{0.6019} & 0.8777  & \cellcolor{lightgreen}{0.9530} \\
		TarDAL & 59.80  & 0.7977  & 0.8993  & 0.4812  & 0.4424  & 0.8651  & 0.9455  \\
		MRFS  & 58.28  & 0.8032  & 0.9026  & 0.5479  & 0.5985  & \cellcolor{lightblue}{0.8826} & \cellcolor{lightblue}{0.9533} \\
		TIMFusion & \cellcolor{lightred}{60.86} & 0.7985  & 0.8998  & 0.5407  & 0.5965  & 0.8799  & 0.9492  \\
		SAGE  & 54.89  & \cellcolor{lightblue}{0.8093} & \cellcolor{lightblue}{0.9066} & \cellcolor{lightgreen}{0.5855} & 0.5921  & \cellcolor{lightgreen}{0.8821} & \cellcolor{lightred}{0.9535} \\
		\hline
		IDF-TDDT & 59.88  & 0.8077  & 0.9049  & 0.5836  & \cellcolor{lightblue}{0.6135} & 0.8812  & 0.9507  \\
		CLDyN & \cellcolor{lightgreen}{60.17} & \cellcolor{lightgreen}{0.8091} & \cellcolor{lightgreen}{0.9055} & \cellcolor{lightblue}{0.5939} & 0.5334  & 0.8783  & 0.9489  \\
		\hline
		Ours  & \cellcolor{lightblue}{60.64} & \cellcolor{lightred}{0.8127} & \cellcolor{lightred}{0.9096} & \cellcolor{lightred}{0.6157} & \cellcolor{lightred}{0.6289} & \cellcolor{lightred}{0.8833} & 0.9506  \\
		\bottomrule
	\end{tabular}\vspace{-3mm}
	\label{tab3}%
\end{table}%

\subsubsection{Comparison with Multi-Task-Aware Methods}

We further compare the proposed method with multi-task-aware methods to verify its task generalization ability in diverse open-task scenarios. Specifically, under the diverse open-task scenario settings, representative multi-task-aware methods, i.e., IDF-TDDT \cite{70} and CLDyN \cite{125}, are trained using the corresponding seen tasks. In the evaluation of both seen and unseen tasks, the fused images generated by these methods are directly deployed to the corresponding task networks. As shown in Fig. \ref{fig6} and Tables \ref{tab1}, \ref{tab2}, and \ref{tab3}, compared with multi-task-aware methods, the proposed method achieves more stable and competitive performance on multiple seen and unseen tasks, and exhibits stronger robustness across the diverse open-task scenarios. In contrast, multi-task-aware methods show certain instability in their generalization ability across different tasks. That is, they can maintain relatively consistent performance on some tasks, but suffer from significant performance degradation on others. For example, for CLDyN on the OD task, regardless of whether OD is involved in training as a seen task, its ${\rm{mA}}{{\rm{P}}_{50 \to 95}}$ remains around 0.6, indicating stable detection performance. However, when Seg is a seen task, its mIoU is about 60.3, whereas when Seg is an unseen task, its mIoU drops to 55.42. This indicates that the method exhibits obvious differences in generalization ability across different tasks, thereby limiting its adaptability in open-task scenarios. As shown in Figs. \ref{fig7}, \ref{fig8}, and \ref{fig9}, the qualitative results demonstrate that the results of the proposed method on multiple seen and unseen tasks are closer to the ground truth (GT). In summary, both quantitative and qualitative results demonstrate the advantages of the proposed method over existing multi-task-aware methods in multi-task adaptability and open-task generalization.

\subsection{Fusion Performance Comparison}

\begin{figure*}
	\begin{center}
		\includegraphics[width=0.90\linewidth]{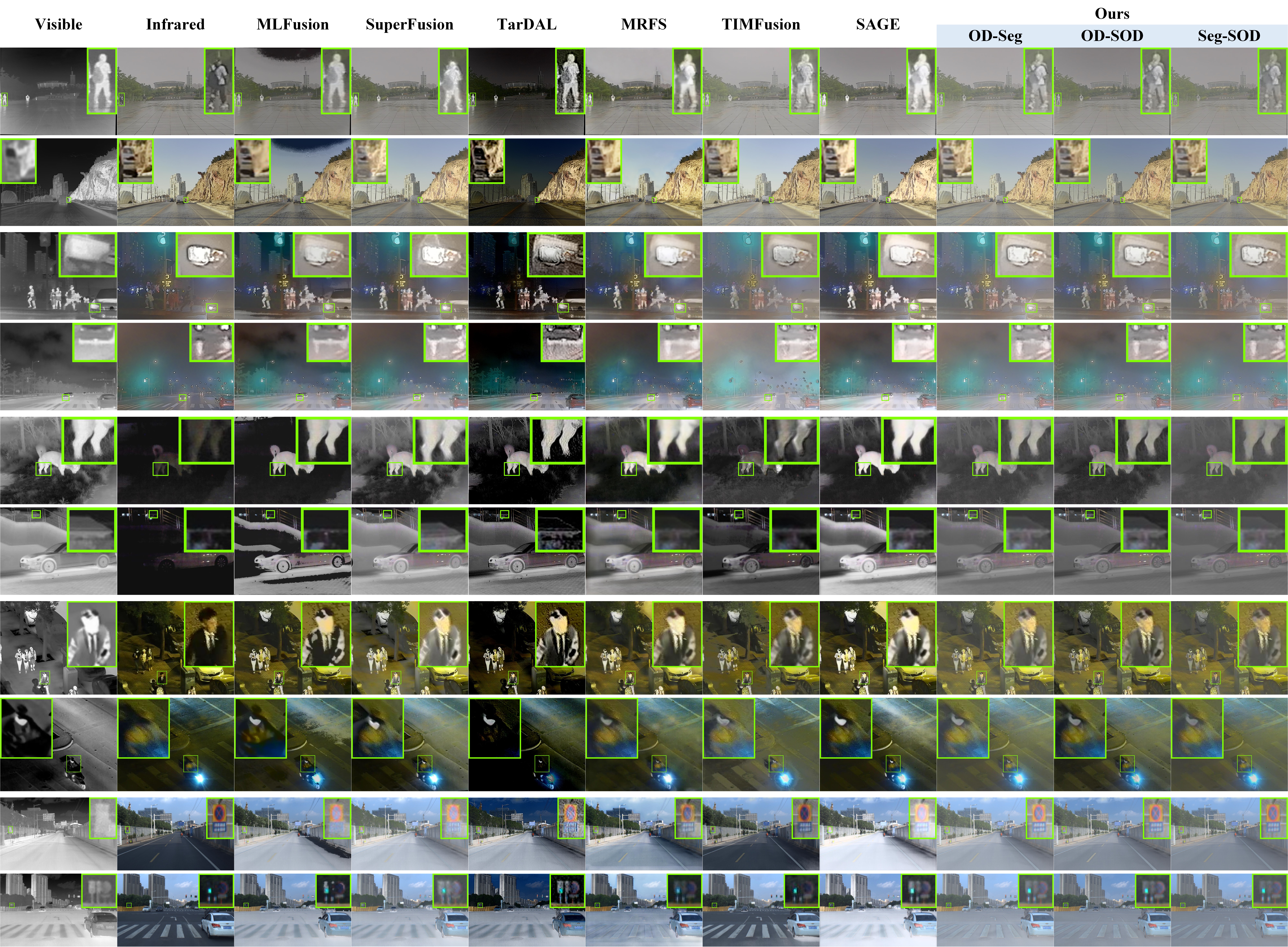}
	\end{center}
	\caption{Qualitative comparison of the proposed method with representative fusion methods on the M$^3$FD, FMB, VT5000, LLVIP, and VTD datasets. The first and second columns show the visible and infrared source images, respectively, and the third to eleventh columns present the fused images of the compared methods. Every two rows correspond to one dataset, showing results from M$^3$FD, FMB, VT5000, LLVIP, and VTD from top to bottom, respectively.}
	\label{fig10}
\end{figure*}

To verify the fusion performance of the proposed method, we comprehensively compare it with representative state-of-the-art fusion methods on five datasets, including MLFusion, SuperFusion, TarDAL, MRFS, TIMFusion, and SAGE. In the fusion performance evaluation, we further analyze the models obtained under three open-task scenario settings. As shown in Table \ref{tab4}, the fused images generated by the proposed method under the three open-task scenarios achieve favorable performance on most evaluation metrics and rank among the top three in most cases. This conclusion is further supported by the visualization results. As shown in Fig. \ref{fig10}, the fused images generated by the proposed method can effectively integrate thermal radiation information from the infrared modality while preserving texture details from the visible modality, thereby enhancing the saliency of target objects concerned by downstream tasks and maintaining favorable overall visual consistency with the visible images. In summary, quantitative and qualitative results demonstrate that the proposed method achieves competitive fusion performance.

\begin{figure*}
	\begin{center}
		\includegraphics[width=0.90\linewidth]{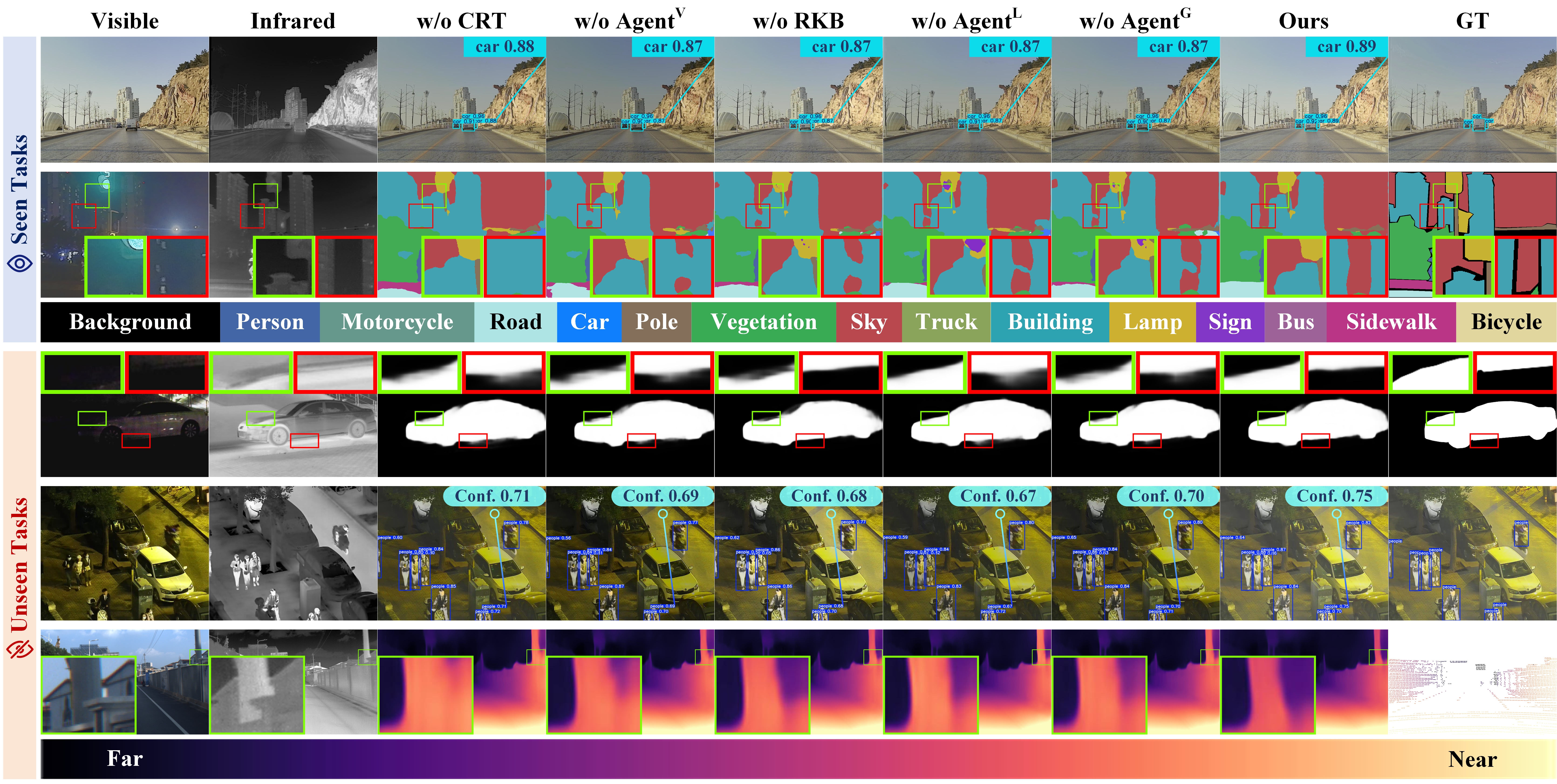}
	\end{center}
	\caption{Qualitative comparison between the complete model and ablation models in the OD-Seg open-task scenario. The first and second columns show the visible and infrared source images, respectively. The third to eighth columns present the downstream task results, and the ninth column shows the Ground Truth (GT). The first to fifth rows correspond to the results of OD, Seg, SOD, PD, and DE, respectively.}
	\label{fig11}
\end{figure*}

\subsection{Ablation Study}

The proposed method is mainly built upon three mechanisms: Causal Requirement Tracing Task Localization, History-Analysis Multi-Objective Balancing, and Task-Level-Correction Conflict Mitigation. The first mechanism consists of the CRT module, ${\rm{Agent}^v}$, and the RKB, while the latter two are centered on ${\rm{Agent}^l}$ and ${\rm{Agent}^g}$, respectively. To verify the effectiveness of these key components, we conduct ablation experiments under the \textbf{OD-Seg open-task setting}.

\textbf{Effectiveness of the CRT module}: The CRT module constructs task-specific causal requirement maps to describe task-specific requirements. To verify the effectiveness of the CRT module, we remove this module and replace the causal requirement map in the state with the features fed back from downstream tasks. This ablation model is denoted as “\textbf{w/o CRT}”. As shown in Table \ref{tab5}, after removing the CRT module, the model exhibits significantly lower performance than the complete model on the unseen downstream tasks, i.e., SOD, PD, and DE, indicating that its generalization ability to unseen tasks is substantially weakened. This conclusion can also be further verified from the qualitative results. As shown in Fig. \ref{fig11}, the results generated by the w/o CRT model on the SOD task suffer from obvious boundary blurring. On the PD task, the detection confidence of pedestrian targets is significantly reduced. On the DE task, the model misclassifies distant buildings as foreground regions, leading to obvious deviations in depth prediction.

\textbf{Effectiveness of ${\bf{Agent}^v}$}: $\mathrm{Agent}^{v}$ adaptively integrates requirement knowledge vectors according to the causal requirement maps, thereby guiding the generation of requirement-customized fused images. To verify the effectiveness of ${\rm{Agen}}{{\rm{t}}^v}$, we remove this component and directly feed the causal requirement maps into TAFN. This ablation model is denoted as “\textbf{w/o ${\rm{Agent}^v}$}”. It should be noted that the RKB is highly correlated with ${\rm{Agent}^v}$ and is therefore also removed accordingly. As shown in Table \ref{tab5} and Fig. \ref{fig11}, after removing ${\rm{Agent}^v}$, the model suffers from significant performance degradation on unseen downstream tasks. This indicates that although the CRT module can generate causal requirement maps reflecting the potential requirements of tasks, without effective interpretation by ${\rm{Agent}^v}$, the model still struggles to fully understand task-specific requirements, thus limiting its generalization ability in open-task scenarios. Furthermore, the main difference between the w/o RKB model and the w/o ${\rm{Agent}^v}$ model lies in whether ${\rm{Agent}^v}$ is retained. Compared with w/o RKB, w/o ${\rm{Agent}^v}$ still exhibits more significant performance degradation on unseen tasks, further demonstrating that ${\rm{Agent}^v}$ is an important component for adapting to the requirements of unseen tasks.

\begin{table*}[htbp]
	\centering
	\caption{Quantitative comparison of the proposed method with representative fusion methods on the M$^3$FD, FMB, VT5000, LLVIP, and VTD datasets. The best, second-best, and third-best metric values are highlighted with {\setlength{\fboxsep}{2.3pt}\colorbox{lightred}{Red}}, {\setlength{\fboxsep}{2.3pt}\colorbox{lightblue}{Blue}}, and {\setlength{\fboxsep}{2.3pt}\colorbox{lightgreen}{Green}} backgrounds, respectively.}

	\renewcommand{\arraystretch}{1.6}
	\setlength{\tabcolsep}{1.7pt}
	\scriptsize

	\begin{tabular}{cc|cccc|cccc|cccc|cccc|cccc}
		\toprule
		\multicolumn{2}{c|}{\multirow{2}{*}{\textbf{Methods}}} & \multicolumn{4}{c|}{\textbf{M$^3$FD}}     & \multicolumn{4}{c|}{\textbf{FMB}}      & \multicolumn{4}{c|}{\textbf{VT5000}}   & \multicolumn{4}{c|}{\textbf{LLVIP}}    & \multicolumn{4}{c}{\textbf{VTD}} \\
		\cline{3-22}    \multicolumn{2}{c|}{} & ${Q_{CB}}$    & SSIM  & CC    & ${N_{AB/F}}$  & ${Q_{CB}}$   & SSIM  & CC    & ${N_{AB/F}}$  & ${Q_{CB}}$   & SSIM  & CC    & ${N_{AB/F}}$  & ${Q_{CB}}$   & SSIM  & CC    & ${N_{AB/F}}$  & ${Q_{CB}}$   & SSIM  & CC    & ${N_{AB/F}}$ \\
		\midrule
		\multicolumn{2}{c|}{MLFusion} & 0.4272  & 1.3874  & 0.6251  & 0.01913  & 0.4313  & 1.4688  & 0.6638  & 0.02264  & 0.4356  & 1.5006  & 0.7014  & 0.02632  & 0.3643  & 1.2497  & \cellcolor{lightgreen}{0.8428} & 0.01702  & 0.3578  & 1.2877  & 0.7398  & 0.04402  \\
		\multicolumn{2}{c|}{SuperFusion} & 0.3979  & 1.3696  & 0.6331  & 0.01251  & 0.4148  & 1.4663  & 0.6745  & 0.00942  & 0.4662  & 1.4610  & 0.7018  & 0.01710  & 0.3463  & 1.2498  & \cellcolor{lightblue}{0.8435} & 0.02642  & 0.3188  & 1.2226  & 0.7671  & 0.03353  \\
		\multicolumn{2}{c|}{TarDAL} & 0.3605  & 0.7739  & 0.5077  & 0.16685  & 0.3367  & 0.6790  & 0.5818  & 0.14852  & 0.4193  & 1.0466  & 0.6396  & 0.20763  & 0.3682  & 0.5994  & 0.7612  & 0.09780  & 0.3631  & 0.7658  & 0.6838  & 0.29679  \\
		\multicolumn{2}{c|}{MRFS} & 0.4218  & \cellcolor{lightgreen}{1.4152} & \cellcolor{lightgreen}{0.6370} & 0.01186  & 0.4329  & \cellcolor{lightgreen}{1.4968} & 0.6701  & 0.01047  & 0.4507  & 1.5232  & 0.6957  & 0.01480  & 0.3695  & 1.3099  & 0.8364  & 0.00249  & 0.3707  & 1.3267  & 0.7570  & 0.01539  \\
		\multicolumn{2}{c|}{TIMFusion} & 0.3405  & 1.3523  & 0.6325  & 0.00753  & 0.3180  & 1.4126  & 0.6532  & 0.00749  & \cellcolor{lightgreen}{0.5139} & 1.4101  & 0.6947  & 0.04872  & 0.3150  & 1.1370  & \cellcolor{lightred}{0.8936} & 0.00522  & 0.3910  & 1.2601  & 0.7262  & 0.02043  \\
		\multicolumn{2}{c|}{SAGE} & \cellcolor{lightblue}{0.4386} & \cellcolor{lightblue}{1.4234} & 0.6173  & 0.01338  & \cellcolor{lightblue}{0.4421} & \cellcolor{lightblue}{1.5033} & 0.6767  & 0.01078  & 0.4604  & \cellcolor{lightgreen}{1.5253} & 0.7054  & 0.01112  & 0.4317  & 1.3259  & 0.8008  & 0.01166  & 0.3424  & 1.3003  & 0.7355  & 0.02842  \\
		\hline
		\multirow{3}[2]{*}{Ours} & OD-Seg & \cellcolor{lightred}{0.4406} & 1.3826  & \cellcolor{lightblue}{0.6588} & \cellcolor{lightblue}{0.00107} & \cellcolor{lightgreen}{0.4409} & 1.4508  & \cellcolor{lightblue}{0.6949} & \cellcolor{lightblue}{0.00070} & \cellcolor{lightblue}{0.5257} & 1.5147  & \cellcolor{lightgreen}{0.7198} & \cellcolor{lightblue}{0.00095} & \cellcolor{lightblue}{0.5257} & \cellcolor{lightgreen}{1.5147} & 0.7198  & \cellcolor{lightblue}{0.00095} & \cellcolor{lightblue}{0.3958} & \cellcolor{lightgreen}{1.3269} & \cellcolor{lightgreen}{0.7989} & \cellcolor{lightblue}{0.00071} \\
		& OD-SOD & \cellcolor{lightgreen}{0.4308} & \cellcolor{lightred}{1.4280} & 0.6198  & \cellcolor{lightgreen}{0.00116} & \cellcolor{lightred}{0.4507} & \cellcolor{lightred}{1.5092} & \cellcolor{lightgreen}{0.6785} & \cellcolor{lightgreen}{0.00077} & 0.5077  & \cellcolor{lightred}{1.5562} & \cellcolor{lightblue}{0.7228} & \cellcolor{lightgreen}{0.00125} & \cellcolor{lightgreen}{0.5077} & \cellcolor{lightred}{1.5562} & 0.7228  & \cellcolor{lightgreen}{0.00125} & \cellcolor{lightgreen}{0.3914} & \cellcolor{lightred}{1.3744} & \cellcolor{lightred}{0.8295} & \cellcolor{lightgreen}{0.00308} \\
		& Seg-SOD & 0.3831  & 1.3998  & \cellcolor{lightred}{0.6696} & \cellcolor{lightred}{0.00050} & 0.3725  & 1.4696  & \cellcolor{lightred}{0.7054} & \cellcolor{lightred}{0.00033} & \cellcolor{lightred}{0.5539} & \cellcolor{lightblue}{1.5356} & \cellcolor{lightred}{0.7285} & \cellcolor{lightred}{0.00031} & \cellcolor{lightred}{0.5539} & \cellcolor{lightblue}{1.5356} & 0.7285  & \cellcolor{lightred}{0.00031} & \cellcolor{lightred}{0.3996} & \cellcolor{lightblue}{1.3358} & \cellcolor{lightblue}{0.8214} & \cellcolor{lightred}{0.00019} \\
		\bottomrule
	\end{tabular}
	\label{tab4}%
\end{table*}%

\textbf{Effectiveness of RKB}: The RKB aims to provide ${\rm{Agent}^v}$ with rich image-level requirement-related knowledge to support its interpretation of the causal requirement maps. To verify the effectiveness of the RKB, we remove this component and enable ${\rm{Agent}^v}$ to directly generate requirement knowledge vectors to guide TAFN in reconstructing task requirement-customized fused images. This ablation model is denoted as “\textbf{w/o RKB}”. As shown in Table \ref{tab5}, after removing the RKB, the model exhibits performance degradation to varying degrees on multiple seen and unseen downstream tasks. In addition, compared with w/o CRT and w/o ${\rm{Agent}^v}$, w/o RKB leads to relatively smaller performance degradation. This indicates that, on the basis of the CRT module and ${\rm{Agent}^v}$, the RKB can provide rich knowledge support for the requirement interpretation process of ${\rm{Agent}^v}$, thereby further improving the generalization ability of the model to unseen tasks. As shown in Fig. \ref{fig11}, compared with the w/o RKB model, the complete model produces prediction results that are closer to the Ground Truth (GT) on different downstream tasks, further verifying the effectiveness of the RKB.

\begin{table}[htbp]
	\centering
	\caption{Quantitative comparison between the complete model and ablation models in the OD-Seg open-task scenario. The best metric values are highlighted with {\setlength{\fboxsep}{2.3pt}\colorbox{lightred}{Red}} backgrounds.}

	\renewcommand{\arraystretch}{1.6}
	\setlength{\tabcolsep}{2pt}
	\scriptsize 
	
	\begin{tabular}{c|c|c|cc|c|cc}
		\toprule
		\multirow{3}{*}{\textbf{Methods}} & \multicolumn{2}{c|}{\textbf{Seen Tasks}} & \multicolumn{5}{c}{\textbf{Unseen Tasks}} \\
		\cline{2-8}          & \textbf{OD}    & \textbf{Seg}   & \multicolumn{2}{c|}{\textbf{SOD}} & \textbf{PD}    & \multicolumn{2}{c}{\textbf{DE}} \\
		\cline{2-8}          & ${\rm{mA}}{{\rm{P}}_{50 \to 95}}$ & mIoU  & ${\rm{m}}{F_\beta }$    & ${E_m}$    & ${\rm{mA}}{{\rm{P}}_{50 \to 95}}$ & $\Delta  < {1.25^2}$    & $\Delta  < {1.25^3}$ \\
		\midrule
		w/o CRT & 0.6104  & 59.38  & 0.8076  & 0.9063  & 0.6032  & 0.8798  & 0.9484  \\
		w/o ${\rm{Agent}^v}$ & 0.6093  & 59.86  & 0.8078  & 0.9058  & 0.6072  & 0.8832  & 0.9522  \\
		w/o RKB & 0.6139  & 59.84  & 0.8093  & 0.9069  & 0.6164  & \cellcolor{lightred}{0.8852} & 0.9532  \\
		w/o ${\rm{Agent}^l}$ & 0.6113  & 59.23  & 0.8096  & 0.9075  & 0.6179  & 0.8827  & 0.9503  \\
		w/o ${\rm{Agent}^g}$ & 0.6182  & 59.56  & 0.8078  & 0.9068  & 0.6212  & 0.8821  & 0.9503  \\
		Ours  & \cellcolor{lightred}{0.6184} & \cellcolor{lightred}{60.89} & \cellcolor{lightred}{0.8137} & \cellcolor{lightred}{0.9103} & \cellcolor{lightred}{0.6304} & 0.8847  & \cellcolor{lightred}{0.9534} \\
		\bottomrule
	\end{tabular}
	\label{tab5}%
\end{table}%

\textbf{Effectiveness of ${\bf{Agent}^l}$}: $\mathrm{Agent}^{l}$ is introduced to adaptively regulate the task-aware loss terms, thereby maintaining balanced optimization across tasks. To verify the effectiveness of ${\rm{Agent}^l}$, we remove this component and adopt fixed weights as balancing parameters in the task-aware loss function. This ablation model is denoted as “\textbf{w/o ${\bf{Agent}^l}$}”. As shown in Table \ref{tab5}, after removing ${\rm{Agent}^l}$, the Seg task suffers from more significant performance degradation than the OD task, indicating that obvious optimization imbalance can easily occur among seen tasks. In addition, the qualitative results in Fig. \ref{fig11} further verify this phenomenon. Compared with the complete model, the prediction results of the w/o ${\rm{Agent}^l}$ model on the Seg task exhibit obvious differences from the Ground Truth (GT), manifested as incomplete semantic region partitioning or incorrect segmentation categories.

\textbf{Effectiveness of ${\bf{Agent}^g}$}: $\mathrm{Agent}^{g}$ is introduced to correct the multi-task joint gradient using task-specific gradients, thereby alleviating gradient conflicts during multi-task optimization. To verify the effectiveness of ${\rm{Agent}^g}$, we remove this component and update the model only with the multi-task joint gradient. This ablation model is denoted as “\textbf{w/o ${\bf{Agent}^g}$}”. As shown in Table \ref{tab5} and Fig. \ref{fig11}, after removing ${\rm{Agent}^g}$, the model tends to be more biased toward detection-related tasks. Specifically, among the seen tasks, the OD performance of the w/o ${\rm{Agent}^g}$ model is relatively close to that of the complete model, while the Seg performance decreases significantly, indicating that without gradient correction, the model struggles to simultaneously satisfy the optimization requirements of different seen tasks. Similarly, among the unseen tasks, the performance degradation on the detection-related PD task is relatively small, whereas the other unseen tasks are affected more substantially.

\subsection{Limitations and Future Work}

Although the proposed method demonstrates favorable task generalization ability in diverse open-task scenarios, its applicability to broader image fusion scenarios still has certain limitations. Specifically, real-world applications involve various types of image pairs to be fused, such as medical images, multi-exposure images, and remote sensing images. However, the proposed method is mainly designed for IR-VIS image fusion. Therefore, without specific optimization for other image types, its fusion performance on the above image fusion tasks may be affected. In future work, we will explore an open-task-aware unified image fusion framework to accommodate multiple image fusion types and their downstream task requirements in real-world open-task scenarios, thereby further enhancing the application potential of image fusion techniques across different domains.

\section{Conclusion}\label{sec5}

This paper proposes an open-task-aware infrared and visible image fusion method, termed the Causal Requirement Tracing-Guided Hierarchical Multi-Agent Regulation framework (CRT-HMAR). Specifically, CRT-HMAR introduces a Causal Requirement Tracing Task Localization mechanism, which constructs causal requirement maps by actively intervening in key image information and observing the response variations of downstream task networks. Based on these maps, a requirement analysis agent adaptively aggregates task-specific requirement knowledge vectors to guide the generation of requirement-customized fused images. Furthermore, History-Analysis Multi-Objective Balancing and Task-Level-Correction Conflict Mitigation mechanisms are introduced to form a hierarchical regulation chain of “requirement interpretation $\to$ task balancing $\to$ conflict mitigation”. Through multiple collaborative agents, CRT-HMAR dynamically regulates open-task requirement modeling, balanced multi-task optimization, and gradient conflict mitigation. Extensive experiments on diverse open-task scenarios involving five downstream tasks demonstrate that the proposed method maintains the performance and balance of seen tasks while achieving stronger generalization to unseen tasks. These results indicate that CRT-HMAR shifts IR-VIS image fusion from task-oriented modeling toward requirement-oriented modeling, promoting its extension from closed-task settings to real-world open-task scenarios.

\bibliographystyle{IEEEtran}
\bibliography{bibliography}

\vfill

\end{document}